\documentclass{article}

\usepackage{microtype}
\usepackage{graphicx}
\usepackage{booktabs} 
\usepackage{url}
\usepackage{soul}
\usepackage{subcaption}
\usepackage{multirow}

\usepackage{hyperref}

\usepackage[preprint]{icml2023}

\usepackage{amsmath}
\usepackage{amssymb}
\usepackage{mathtools}
\usepackage{amsthm}

\usepackage{amsmath,amsfonts,bm}

\def\eqref#1{equation~\ref{#1}}

\def\1{\bm{1}}

\def\vone{{\bm{1}}}

\def\vm{{\bm{m}}}

\def\mM{{\bm{M}}}

\def\mW{{\bm{W}}}

\DeclareMathAlphabet{\mathsfit}{\encodingdefault}{\sfdefault}{m}{sl}
\SetMathAlphabet{\mathsfit}{bold}{\encodingdefault}{\sfdefault}{bx}{n}

\def\sM{{\mathbb{M}}}

\def\sR{{\mathbb{R}}}

\usepackage[capitalize,noabbrev]{cleveref}

\theoremstyle{plain}

\theoremstyle{definition}

\theoremstyle{remark}

\usepackage{color}
\usepackage{xcolor}
\usepackage{colortbl}
\definecolor{lgreen}{HTML}{5fcf5f} 
\definecolor{lred}{HTML}{db6063} 
\definecolor{amazonite}{RGB}{68, 215, 182}

\icmltitlerunning{Structured Pruning Adapters}

\begin{document}

\twocolumn[
\icmltitle{Structured Pruning Adapters}



\icmlsetsymbol{equal}{*}

\begin{icmlauthorlist}
\icmlauthor{Lukas Hedegaard}{au}
\icmlauthor{Aman Alok}{cactus}
\icmlauthor{Juby Jose}{cactus}
\icmlauthor{Alexandros Iosifidis}{au}
\end{icmlauthorlist}

\icmlaffiliation{au}{Department of Electrical and Computer Engineering, Aarhus University, Aarhus, Denmark}
\icmlaffiliation{cactus}{Cactus Communications}

\icmlcorrespondingauthor{Lukas Hedegaard}{lhm@ece.au.dk}

\icmlkeywords{Machine Learning, ICML, Pruning, Adapters, Transfer Learning}

\vskip 0.3in
]



\printAffiliationsAndNotice{}  

\begin{abstract}

Adapters are a parameter-efficient alternative to fine-tuning, which augment a frozen base network to learn new tasks.
Yet, the inference of the adapted model is often slower than the corresponding fine-tuned model.
To improve on this, we propose Structured Pruning Adapters (SPAs), a family of compressing, task-switching network adapters, that accelerate and specialize networks using tiny parameter sets and structured pruning.
Specifically, we propose a channel-based SPA and evaluate it with a suite of pruning methods on multiple computer vision benchmarks.
Compared to regular structured pruning with fine-tuning, our channel-SPAs improve accuracy by 6.9$\%$ on average while using half the parameters at 90\% pruned weights. Alternatively, they can learn adaptations with 17$\times$ fewer parameters at 70\% pruning with 1.6$\%$ lower accuracy.
Our experimental code and Python library of adapters are available at 
\url{github.com/lukashedegaard/structured-pruning-adapters}.

\end{abstract}

\section{Introduction}
Fine-tuning is an established approach to parameter-based transfer learning from a source model pre-trained on a large dataset to a target task with limited training data. 
However, the resulting model retains the same parameter count and computational characteristics as the source model, even when solving a considerably simpler task. 
A group of \textit{fine-pruning} methods~\cite{li2017pruning, molchanov2017prining, sun17meprop, yeom2021pruning, sanh2020movement, lagunas2021block} have combined pruning with fine-tuning to reduce model size while learning parameters for the target task. Generally, this results in compressed models that retain performance down to at least half the relative weight density and which may be better suited for resource-constrained deployments, such as mobile devices, robotics applications, and settings necessitating low-latency predictions. 

Meanwhile, Adapters~\cite{rebuffi2017learning, rebuffi2018efficient} have emerged as a viable alternative to fine-tuning for multi-domain deep neural networks (DNNs), where a single source DNN is specialized and sequentially used for multiple tasks. Instead of continuing training of the source DNN weights directly, Adapters introduce parameter-efficient layer add-ons, which are trained instead. As these add-ons are much more compact than the source DNN weights, they can be transmitted and stored at low cost. This is very useful, \textit{e.g.}, for edge devices and federated learning~\cite{mcmahan2016communication}.
However, prior work has largely ignored the computational efficiency aspects of Adapters, which either increase the complexity of the network~\cite{he2022towards, zhu2021counter, liliang2021prefix, houlsby2019parameter, pfeiffer2021adapterfusion, mahabadi2021compacter} or leave it unaltered, at best, by utilizing structures that can be fused with the original weights~\cite{rebuffi2017learning, rebuffi2018efficient, hu2022lora}.

In general, a deployed model must strike a balance between predictive performance, storage requirements, inference speed, and flexibility of use. 
%
While the combination of pruning and fine-tuning can produce compressed models with good performance at an order of magnitude fewer parameters compared to the source model, \textit{Structured Pruning Adapters (SPAs)} can improve upon this by another order of magnitude for task-switching networks.
%
%
%
Specifically, we propose the \textit{Structured Pruning Low-rank Adapter} (SPLoRA) for channel-based pruning and compare its performance and parameter count to pruning with fine-tuning in weight-based transfer learning from ResNet weights pretrained on ILSVRC 2012~\cite{russakovsky2015imagenet} to the image classification benchmarks CIFAR-10~\cite{krizhevsky09learning}, Oxford Flowers 102~\cite{nilsback2008automated}, and Cats and Dogs~\cite{elson2007asirra}. Considering four different pruning methods, we find that SPLoRA not only reduces parameter requirements per task massively, but also retains predictive accuracy better than fine-tuning under aggressive pruning. 
%

In the remainder of this paper, we describe related work on Adapters (\cref{sec:adapters}) and Pruning (\cref{sec:pruning}), a framework for SPAs (\cref{sec:spas}), the SPLoRA adapter (\cref{sec:splora}), 
experimental comparisons with fine-pruning (\cref{sec:experiments}), and conclusions (\cref{sec:conclusion}).
\section{Related Work}

\subsection{Adapter methods}\label{sec:adapters}
When multiple specialized versions of a network are deployed on the same device and storage requirements are strict, 
Adapters~\cite{rebuffi2017learning} provide a low-parameter-count alternative to fine-tuning. Instead of deploying multiple sets of full network weights, a single set of full weights can be deployed alongside multiple adapter weights, which augment the main network.
For Convolutional Neural Networks (CNNs), point-wise convolutions can be introduced in series~\cite{rebuffi2017learning} or parallel~\cite{rebuffi2018efficient} with a residual connection to adapt fixed source weights to new tasks. 
For Transformer-based networks, prior work explored bottleneck projections with~\cite{zhu2021counter} and without~\cite{hu2022lora} low-dimensional non-linearity in parallel with the fixed dense layers of the original network. Prefix-tuning~\cite{liliang2021prefix}, which learns task-specific prompt tokens to adapt the network, may be considered a parallel adapter as well~\cite{he2022towards}. Adapter blocks can also be interspersed in series with existing layers~\cite{houlsby2019parameter, pfeiffer2021adapterfusion, mahabadi2021compacter}. 
He at al.~\cite{he2022towards} proposed a combination of the above. Finally, several works~\cite{stickland2019bertandpals, pfeiffer2021adapterfusion, ruckle2021adapterdrop} explored the use of multi-task adapters.


While the above-described methods succeed in learning parameter-efficient network add-ons with very small storage requirements, these adapters often incur an additional computational cost beyond the original network.
Considering that the adapted target tasks are often simpler than the source task, it is reasonable to assume that a derived network adaptation can be learned, which reduces computational complexity as well.

\subsection{Efficiency approaches}\label{sec:efficiency-approaches}

Multiple approaches have been proposed to reduce the compute and memory footprint of neural networks.
\textit{Knowledge distillation}\footnote{Knowledge distillation~\cite{hinton15distilling} using the unpruned model as the teacher has been found to help pruning methods~\cite{sanh2020movement, lagunas2021block} retain accuracy better using fine-pruning. We expect this to be true for SPAs as well, but we leave the validation to future work as none of the considered fine-pruning baselines employed knowledge distillation.
}~\cite{hinton15distilling, gou2021knowledge} utilize a large network as a teacher for a smaller network, which has more desirable memory and computational characteristics.
\textit{Efficient architectures}~\cite{tan2019efficientnet, feichtenhofer2020x3d, hoffmann2022chinchilla} define and optimize expressive yet efficient architectural blocks from random initialization under a multi-metric optimization goal. 
Low-rank factorizations~\cite{tran2018improving,guo2019compressing} approximate large tensor weights by factorizing them into multiple lower-rank weights.
\textit{Continual Inference Networks}~\cite{hedegaard2022co3d, hedegaard2022cotrans} reuse the network weights of prior DNNs with a temporal component and accelerate them for online stream processing via optimized computational sequences and appropriate intra-layer caching.
\textit{Quantization approaches}~\cite{gray1998quantization, liang2021pruning} reduce model size and run-time costs via low-resolution numerical representations of network weights.
Finally, \textit{Pruning methods}~\cite{lecun1989optimal, han2015learning, frankle2019lottery} entirely remove unnecessary network weight from a pre-trained model.
While all of these are interesting research avenues both in isolation and combination, we focus on pruning-methods hereafter.

\subsection{Pruning methods}\label{sec:pruning}
DNNs can be pruned at multiple levels: \textit{Unstructured} pruning of individual weights results in sparse weight matrices which reduce parameter count but require specialized hardware to improve inference characteristics due to the performance disadvantages of sparse matrix multiplication kernels on graphical processingn units (GPUs). 
On the other hand, \textit{structured} pruning approaches, such as the pruning of entire channels~\cite{yeom2021pruning} or blocks~\cite{lagunas2021block} of networks weights, generally provide inference speedup across computational devices~\cite{gray2017gpu}.


Many studies have proposed criteria on \textit{what} to prune.
Early methods~\cite{lecun1989optimal, hassibi1992second} proposed the use of second-order Taylor expansion of the loss Hessian for weight selection. 
As computing the inverse of the Hessian may be computationally intractable, another approach uses a first-order Taylor approximation of the loss change due to the pruning of units~\cite{molchanov2017prining}. A recent work uses fast Hessian-vector products to retain low complexity~\cite{nonnenmacher2022sosp}.
Similarly, the gradient of a weight with respect to the loss can be used for pruning selection~\cite{liu2019channel, sun17meprop, sanh2020movement}. 
Among the simplest approaches is the use of weight magnitudes in pruning selection. In our experimental comparisons, we employ Magnitude Pruning~\cite{han2015learning}, which uses a simple average of weight magnitudes, and Weight Pruning~\cite{li2017pruning}, which uses the $L_p$-norm of weights, for channel selection.
Yeom et al.~\cite{yeom2021pruning} proposed an explainability-inspired approach, computing the relevance of each network component by means of Layer-wise Relevance Pruning (LRP)~\cite{bach2015lrp, montavon2019explainable}. We use this method in \cref{sec:experiments-channel}.

For all the above methods assigning pruning scores, another consideration is whether to rank and select structural units locally within a layer (keeping pruning evenly spread throughout the network) or globally, with a contest among all network layers. We utilize global selection in our experiments (\cref{sec:experiments-channel}).

Beyond the pruning criterion, multiple studies proposed \textit{when} to prune.
One popular approach~\cite{molchanov2017prining, liu2019channel, yeom2021pruning} is to use an iterative pruning and fine-tuning schedule, masking a predefined fraction of units at a time. Alternatively, Automated Gradual Pruning~\cite{zhu2018toprune} allows all weights to be updated throughout the pruning schedule, enabling prior masking choices to be altered. 
We use the former in \cref{sec:experiments-channel}.

\subsection{Transfer pruning}

Pruning is useful not only for compressing a model while retaining predictive performance, but also for transfer learning~\cite{yeom2021pruning, sanh2020movement, lagunas2021block}.
In fact, a task can be ``learned'' simply by selecting the appropriate subset of weights in a network~\cite{mallya2018piggyback, ramanujan2020whats}.

Consider a large (pre-trained) source model $f_s$ and a set of $T$ target tasks for which we desire specialized target models $f_t, t \in \{1..T\}$.
Under the framework of transfer learning with pruning (``transfer pruning''), we can concurrently update and mask weights from a source model to benefit a target task $t$. 
Consider $g: \mW_s \times \Delta\mW_t \times \sM_t \rightarrow \mW_t$, a function that generates target model weights $\mW_t$, given learned update weights $\Delta\mW_t$, source weights $\mW_s$, and a learned masking set $\sM_t$ of retained weight indices.
Given available source weights $\mW_s$, every task-specific model $f_t$ can be stored as the parameters $\Phi_t = \{\Delta\mW_t, \sM_t\}$. 

Under \textit{fine-pruning}~\cite{sanh2020movement}, i.e., concurrent pruning and fine-tuning, $g$ constitutes a direct assignment of weights, $\mW_t \coloneqq g(\mW_s, \Delta\mW_t, \sM_t) = \Delta\mW_t$, where update weights are learned based on a pruned subset, $\{ \mW_s^{(i)}, i \in \sM_t\}$.
Here, the parameters of the task-specific model are $\Phi_t = \{\mW_t, \sM_t\}$, and the size of the target weights is determined by the weight density $d \in (0,1]$ and the size of source weights: 
$\lVert \mW_t \rVert_0 = d \lVert \mW_s \rVert_0$.

\begin{figure}[tb]
    \centering
    \includegraphics[width=0.8\linewidth]{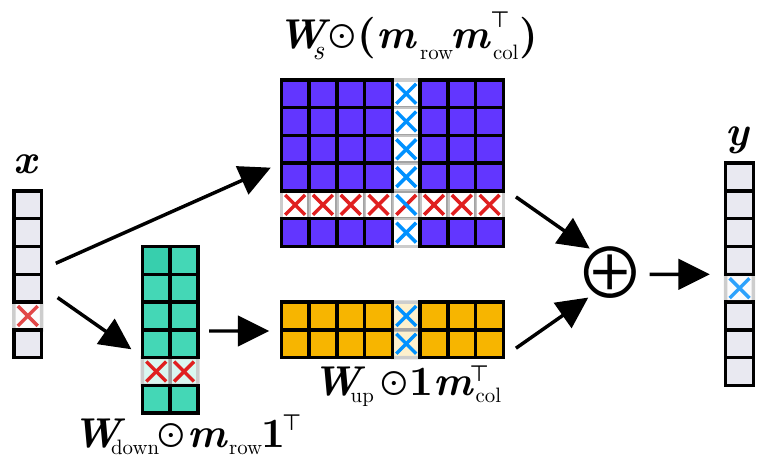}
    \caption{Structured Pruning Low-rank Adapter (SPLoRA). Pruning of in/out channels affects the adapter as well as source weights.}
    \label{fig:splora}
\end{figure}

\section{Structured Pruning Adapters}\label{sec:spas}
Although fine-pruning can successfully produce smaller target weights, i.e., $\lVert \mW_t \rVert_0 < \lVert \mW_s \rVert_0$, the set of weights for all tasks $\{\mW_t\}$ may still consume an intractable amount of storage if many tasks $T$ are involved and/or the average density $\Bar{d}$ is large due to high predictive performance requirements.
Instead, we seek to utilize adapters alongside pruning to produce an extremely compressed parameter set.

Consider the concurrent pruning and adaptation of a frozen source projection matrix $\mW_s \in \sR^{n \times m}$ with an index mask $\mM \in \{0,1\}^{n \times m}$ and an adapter function $a$. 
While applicable adapters have been extensively studied (see \cref{sec:adapters}), we restrict ourselves to fusible parallel adapters to minimize the run-time of the resulting model. Accordingly, pruning adapters take the following basic form:
\begin{equation}
    \mW_t = (\mW_s + a(\Delta\mW_t)) \odot \mM.
    \label{eq:basic-parallel}
\end{equation}

\subsection{Structured Pruning Low-rank Adapter}
\label{sec:splora}

Unlike unstructed pruning, \textit{structured} methods remove groups of weights and increase computational efficiency.
\textit{Channel} pruning, in particular, maps a dense source matrix to a dense pruned matrix with computational improvements proportional to the number of removed parameters.
A mask $\mM$ in this case can be decomposed as row and column masks
$\vm_\text{row} \in \{0,1\}^{n\times1}$ and $\vm_\text{col}  \in \{0,1\}^{m\times1}$, respectively.
Then, \cref{eq:basic-parallel} can be expressed as follows:
\begin{equation}
    \mW_t = (\mW_s + a(\Delta\mW_t)) \odot \vm_\text{row} \vm_\text{col}^\top.
    \label{eq:channel-parallel}
\end{equation}

A simple realization of a fusible parallel adapter is the Low-rank Adapter (LoRA)~\cite{hu2022lora}:
\begin{equation}
    \mW_t = \mW_s + \mW_\text{down}\mW_\text{up},
    \label{eq:mlora}
\end{equation}
%
%
%
where $\mW_\text{down} \in \sR^{n \times r}$ and $\mW_\text{up} \in \sR^{r \times m}$ are adapter weights and $r$ is the rank hyper-parameter.
Following the derivation in \cref{apx:smola-derivation}, we utilize Equations \ref{eq:channel-parallel} and \ref{eq:mlora} to define the \textbf{S}tructured \textbf{P}runing \textbf{Lo}w \textbf{R}ank \textbf{A}dapter (SPLoRA): 
\begin{equation}
    \mW_t = \mW_s \odot \vm_\text{row} \vm_\text{col}^\top
        + (\mW_\text{down} \odot \vm_\text{row} \vone^\top ) (\mW_\text{up} \odot \vone \vm_\text{col}^\top ).
    \label{eq:splora}
\end{equation}
In this form, we see that channel-pruning affects not only the source weights $\mW_\text{s}$, but also the adapter parameters $\mW_\text{up}$ and $\mW_\text{down}$. This effect is illustrated in \cref{fig:splora}.

Adapters should generally be initialized to perform a near-zero mapping to avoid destabilizing gradients during initial training. Accordingly, we use the initialization $\mW_\text{up}^{(i,j)} \sim \mathcal{U}(-10^{-4}, 10^{-4})$, although other near-zero initialization choices may be equally successful. Experiments validating this choice are presented in \cref{sec:splora-sensitivity}.

\begin{figure}[tb]
    \centering
    \includegraphics[width=0.95\linewidth]{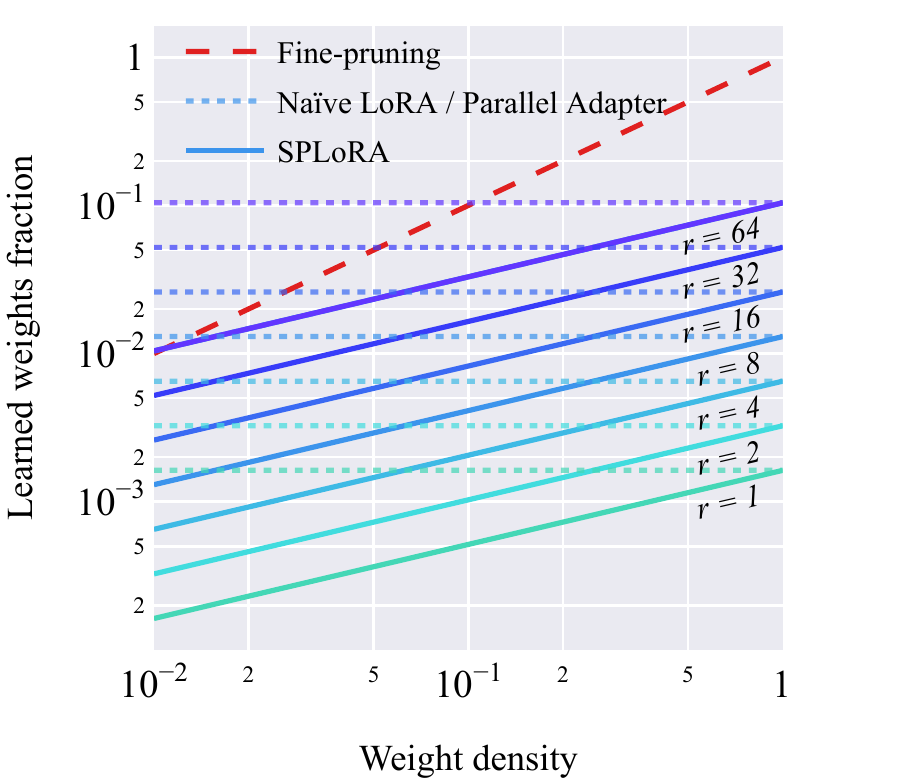}
    \caption{Learned weight fraction ($\lVert \Delta\mW_t \rVert_0 / \lVert \mW_s \rVert_0$) versus weight density ($\lVert \mW_t \rVert_0 / \lVert \mW_s \rVert_0$) 
    for \mbox{$768 \times 3072$} weights under channel-pruning.
    }
    \label{fig:splora-tradeoff-theory}
\end{figure}

For a fine-pruned model weight, $\mW_t$, the updated parameter count is 
$\lVert \vm_\text{row} \rVert_0 \lVert \vm_\text{col} \rVert_0$.
The corresponding SPLoRA has $r (\lVert \vm_\text{row} \rVert_0 + \lVert \vm_\text{col} \rVert_0)$ parameters. 
Parameter comparisons depends on the weight retention fraction, weight matrix shape, and SPLoRA bottleneck dimension $r$. \Cref{fig:splora-tradeoff-theory} visualizes the parameters achieved by varying the hyper-parameters for a $768 \times 3072$ matrix. 
Here, we depict the learned parameter count of fine-pruning and SPLoRA alongside the naïve application of LoRA or, equivalently, the Parallel Adapter~\cite{he2022towards, zhu2021counter} for reference.
While naïve adapter usage requires far-fewer parameters than fine-tuning at high weight density, the parameter reduction achieved by fine-pruning can produce equally few parameters at low weight densities. SPLoRA combines the  benefits of both approaches to produce parameter sets far more compact than those produced by either.

\subsection{Limitations}\label{sec:limitations}
While each set of SPA weights has significantly fewer learned parameters than each set of fine-pruning weights, the superiority of SPAs is dependent on the deployment at hand.
When only a single model is deployed, the two approaches are equivalent in terms of model size, considering that the fused parameter count of adapters is identical to that of the fine-pruned network. 
For $T$ models deployed on one device, the adapter format saves space when 
$T > 1 / \overline{d}$, assuming available source weights $\mW_s$.
Thus, the feasibility depends on the average density $\overline{d}$, with more aggressive pruning requiring more deployed models for SPA formats to consume less storage space.

\subsection{Pruning of adapter-specific parameters}
The design choices made for SPLoRA (\cref{sec:splora}) let us utilize prior channel-based pruning without modification by pruning fused weights, which are identical in form to the source weights, and subsequently pruning the adapter-weights with the same mask. 
The adapters themselves could also be subject to pruning~\cite{ruckle2021adapterdrop}. For SPLoRA, the bottleneck contains $r$ channels that could be pruned per layer to produce a heterogeneous set of layer adapters with different ranks per layer.
%
However, such pruning adds an additional layer of complexity and hyper-parameters to the already complicated pruning methods, which may reduce the attractiveness of adapters. 
Moreover, the pruning of intra-adapter channels only reduces the parameter count and does not yield computational benefits.
Therefore, we limit the pruning in subsequent experiments to the approach we believe has the best usability in practice; we refrain from pruning over internal adapter structures.
\section{Experiments}\label{sec:experiments}\label{sec:experiments-channel}
We seek to compare structured pruning with fine-tuning (fine-pruning) to our proposed channel-based Structured Pruning Adapter method, SPLoRA. 
As both have identical acceleration benefits during inference, our experimental comparison focuses on prediction performance and the number of learned parameters ($\Delta$Params).


\begin{table*}[t]
\caption{Channel-based transfer-pruning from ResNet-50 pre-trained on ImageNet to Cats and Dogs, Oxford Flowers 102, and CIFAR-10 using methods based on Weight~\cite{li2017pruning}, Gradient~\cite{sun17meprop}, Taylor~\cite{molchanov2017prining}, and LRP~\cite{yeom2021pruning}. 
Please note that SPLoRA and LoRA are identical at 100\% density. 
$\Delta$Params and floating point operations (FLOPs) are shown for CIFAR-10. Mean $\pm$ standard deviation is shown for each metric.
Changes specified relative to closest fine-pruning density with same pruning method. Best metric per pruning-method is highlighted with bold.
}
\label{tab:resnet50}
\begin{center}
\resizebox{\textwidth}{!}{
\begin{tabular}{llrlllll}
\toprule 
{\bf Pruning}
    &\multirow{1}{*}{\bf Learning}
    &\multirow{1}{*}{\bf Density}
    &\multicolumn{1}{c}{\bf $\Delta$Params}
    &\multicolumn{1}{c}{\bf FLOPs}
    &\multicolumn{1}{c}{\bf CIFAR-10}
    &\multicolumn{1}{c}{\bf Oxford Flowers 102}
    &\multicolumn{1}{c}{\bf Cats \& Dogs}
    \\
{\bf method}
    & {\bf method}
    &
    & \multicolumn{1}{c}{(K)} 
    & \multicolumn{1}{c}{(M)} 
    & \multicolumn{1}{c}{Acc. (\%)}
    & \multicolumn{1}{c}{Acc. (\%)}
    & \multicolumn{1}{c}{Acc. (\%)}
    \\

\midrule 

& \multirow{1}{*}{Fine-tuning}  & 100\% & $23{,}520.8	$\tiny$\pm 0.0$	  & $1{,}304.7	$\tiny$\pm 0.0$	& $\mathbf{97.10}	$\tiny$\pm 0.12$  & $\mathbf{92.20}	$\tiny$\pm 0.00$ & $\mathbf{99.30}	$\tiny$\pm 0.02$\vspace{3pt}\\
\multirow{1}{*}{Unpruned}           & \multirow{1}{*}{(SP)LoRA-$r32$}   & 100\% & $1{,}644.5 	$\tiny$\pm 0.0$ \textcolor{lgreen}{\small \ ($\downarrow14.3\times$)}	& $1{,}304.7	$\tiny$\pm 0.0$	& $95.32	$\tiny$\pm 0.13$ \textcolor{lred}{\small \ ($-1.8$)} & $78.57	$\tiny$\pm 5.93$ \textcolor{lred}{\small \ ($-13.6$)}  & $98.60	$\tiny$\pm 0.43$\textcolor{lred}{\small \ ($-0.5$)}\vspace{3pt}\\
                                    & \multirow{1}{*}{(SP)LoRA-$r8$}    & 100\% & \phantom{$0{,}$}$\mathbf{466.3}	$\tiny$\pm 0.0$ \textcolor{lgreen}{\small \ ($\downarrow50.4\times$)}	& $1{,}304.7	$\tiny$\pm 0.0$	& $95.35	$\tiny$\pm 0.10$ \textcolor{lred}{\small \ ($-1.8$)} & $80.96	    $\tiny$\pm 5.83$ \textcolor{lred}{\small \ ($-11.2$)} & $98.84	$\tiny$\pm 0.52$\textcolor{lred}{\small \ ($-0.46$)}\\
\midrule 
                                    & \multirow{2}{*}{Fine-pruning} & 30\% & $4{,}427.1                                         $\tiny$\pm 72.5$  & \phantom{$0{,}$}$785.4   $\tiny$\pm 5.4$  & $\mathbf{96.38}                               $\tiny$\pm 0.12$  & $93.64                               $\tiny$\pm 2.15$  & $\mathbf{98.64}                               $\tiny$\pm 0.04$ \\
                                    &                               & 10\% & \phantom{$0{,}$}$599.1                                         $\tiny$\pm 20.1$  & \phantom{$0{,}$}$352.1   $\tiny$\pm 3.9$  & $87.23                               $\tiny$\pm 2.00$  & $72.83                               $\tiny$\pm 0.93$  & $95.42                               $\tiny$\pm 0.31$\vspace{3pt}\\
\multirow{2}{*}{Weight}           
                                    & \multirow{2}{*}{SPLoRA-$r32$} & 30\% & \phantom{$0{,}$}$618.3   $\tiny$\pm 2.5$ \textcolor{lgreen}{\small \ ($\downarrow 7.2\times$)}  & \phantom{$0{,}$}$778.6   $\tiny$\pm 2.9$  & $95.59  $\tiny$\pm 0.22$ \textcolor{lred}{\small \ ($-0.8$)}  & $\mathbf{94.57}   $\tiny$\pm 0.58$ \textcolor{lgreen}{\small \ ($+0.9$)}  & $98.43  $\tiny$\pm 0.13$ \textcolor{lred}{\small \ ($-0.2$)}\\
                                    &                               & 10\% & \phantom{$0{,}$}$294.8   $\tiny$\pm 0.4$ \textcolor{lgreen}{\small \ ($\downarrow 2.0\times$)}  & \phantom{$0{,}$}$\mathbf{335.4}   $\tiny$\pm 7.3$  & $93.04   $\tiny$\pm 0.37$ \textcolor{lgreen}{\small \ ($+5.8$)}  & $89.36  $\tiny$\pm 1.09$ \textcolor{lgreen}{\small \ ($+16.5$)}  & $96.25   $\tiny$\pm 0.99$ \textcolor{lgreen}{\small \ ($+0.8$)}\vspace{3pt}\\
                                    & \multirow{2}{*}{SPLoRA-$r8$}  & 30\% & \phantom{$0{,}$}$210.0  $\tiny$\pm 1.9$ \textcolor{lgreen}{\small \ ($\downarrow 21.1\times$)}  & \phantom{$0{,}$}$773.4   $\tiny$\pm 2.1$  & $94.91  $\tiny$\pm 0.24$ \textcolor{lred}{\small \ ($-1.5$)}  & $92.13  $\tiny$\pm 0.15$ \textcolor{lred}{\small \ ($-1.5$)}  & $98.40  $\tiny$\pm 0.12$ \textcolor{lred}{\small \ ($-0.2$)}\\
                                    &                               & 10\% & \phantom{$0{,}$}$\mathbf{128.9}   $\tiny$\pm 0.1$ \textcolor{lgreen}{\small \ ($\downarrow 4.6\times$)}  & \phantom{$0{,}$}$338.9   $\tiny$\pm 7.0$  & $91.24   $\tiny$\pm 0.51$ \textcolor{lgreen}{\small \ ($+4.0$)}  & $86.49  $\tiny$\pm 0.74$ \textcolor{lgreen}{\small \ ($+13.7$)}  & $96.07   $\tiny$\pm 0.69$ \textcolor{lgreen}{\small \ ($+0.6$)}\\
\midrule
                                    & \multirow{2}{*}{Fine-pruning} & 30\% & $3{,}719.8                                         $\tiny$\pm 59.2$  & \phantom{$0{,}$}$571.9   $\tiny$\pm 6.5$  & $\mathbf{95.95}                               $\tiny$\pm 0.09$  & $\mathbf{94.21}                               $\tiny$\pm 0.74$  & $\mathbf{98.22}                               $\tiny$\pm 0.00$\\
                                    &                               & 10\% & \phantom{$0{,}$}$615.7                                          $\tiny$\pm 4.3$  & \phantom{$0{,}$}$244.7   $\tiny$\pm 3.3$  & $91.83                               $\tiny$\pm 1.17$  & $73.06                               $\tiny$\pm 0.70$  & $95.84                               $\tiny$\pm 0.20$\vspace{3pt}\\
\multirow{2}{*}{Gradient}          
                                    & \multirow{2}{*}{SPLoRA-$r32$} & 30\% & \phantom{$0{,}$}$601.0   $\tiny$\pm 0.4$ \textcolor{lgreen}{\small \ ($\downarrow 6.2\times$)}  & \phantom{$0{,}$}$564.6   $\tiny$\pm 1.5$  & $94.91  $\tiny$\pm 0.09$ \textcolor{lred}{\small \ ($-1.0$)}  & $93.58  $\tiny$\pm 0.43$ \textcolor{lred}{\small \ ($-0.6$)}  & $98.17  $\tiny$\pm 0.08$ \textcolor{lred}{\small \ ($-0.0$)}\\
                                    &                               & 10\% & \phantom{$0{,}$}$293.1   $\tiny$\pm 0.4$ \textcolor{lgreen}{\small \ ($\downarrow 2.1\times$)}  & \phantom{$0{,}$}$\mathbf{244.0}   $\tiny$\pm 1.9$  & $93.65   $\tiny$\pm 0.36$ \textcolor{lgreen}{\small \ ($+1.8$)}  & $91.35  $\tiny$\pm 0.47$ \textcolor{lgreen}{\small \ ($+18.3$)}  & $97.54   $\tiny$\pm 0.17$ \textcolor{lgreen}{\small \ ($+1.7$)}\vspace{3pt}\\
                                    & \multirow{2}{*}{SPLoRA-$r8$}  & 30\% & \phantom{$0{,}$}$205.2  $\tiny$\pm 0.2$ \textcolor{lgreen}{\small \ ($\downarrow 18.1\times$)}  & \phantom{$0{,}$}$565.2   $\tiny$\pm 4.1$  & $94.09  $\tiny$\pm 0.28$ \textcolor{lred}{\small \ ($-1.9$)}  & $91.60  $\tiny$\pm 0.30$ \textcolor{lred}{\small \ ($-2.6$)}  & $98.15  $\tiny$\pm 0.07$ \textcolor{lred}{\small \ ($-0.1$)}\\
                                    &                               & 10\% & \phantom{$0{,}$}$\mathbf{128.3}   $\tiny$\pm 0.0$ \textcolor{lgreen}{\small \ ($\downarrow 4.8\times$)}  & \phantom{$0{,}$}$245.4   $\tiny$\pm 4.0$  & $91.25  $\tiny$\pm 0.20$ \textcolor{lred}{\small \ ($-0.6$)}  & $87.46  $\tiny$\pm 0.71$ \textcolor{lgreen}{\small \ ($+14.4$)}  & $97.19   $\tiny$\pm 0.28$ \textcolor{lgreen}{\small \ ($+1.4$)}\\
\midrule
                                    & \multirow{2}{*}{Fine-pruning} & 30\% & $3{,}392.8                                         $\tiny$\pm 81.1$  & \phantom{$0{,}$}$559.9   $\tiny$\pm 0.7$  & $\mathbf{95.71}                               $\tiny$\pm 0.02$  & $92.91                               $\tiny$\pm 0.56$  & $\mathbf{98.22}                               $\tiny$\pm 0.18$\\
                                    &                               & 10\% & \phantom{$0{,}$}$576.8                                          $\tiny$\pm 9.9$  & \phantom{$0{,}$}$236.9   $\tiny$\pm 3.3$  & $88.07                               $\tiny$\pm 0.66$  & $65.67                               $\tiny$\pm 4.12$  & $95.30                               $\tiny$\pm 0.21$\vspace{3pt}\\
\multirow{2}{*}{Taylor}            
                                    & \multirow{2}{*}{SPLoRA-$r32$} & 30\% & \phantom{$0{,}$}$599.7   $\tiny$\pm 0.9$ \textcolor{lgreen}{\small \ ($\downarrow 5.7\times$)}  & \phantom{$0{,}$}$555.5   $\tiny$\pm 6.7$  & $94.88  $\tiny$\pm 0.21$ \textcolor{lred}{\small \ ($-0.8$)}  & $\mathbf{93.41}   $\tiny$\pm 0.06$ \textcolor{lgreen}{\small \ ($+0.5$)}  & $97.84  $\tiny$\pm 0.48$ \textcolor{lred}{\small \ ($-0.4$)}\\
                                    &                               & 10\% & \phantom{$0{,}$}$292.6   $\tiny$\pm 0.5$ \textcolor{lgreen}{\small \ ($\downarrow 2.0\times$)}  & \phantom{$0{,}$}$\mathbf{242.0}   $\tiny$\pm 1.5$  & $93.27   $\tiny$\pm 0.12$ \textcolor{lgreen}{\small \ ($+5.2$)}  & $91.30  $\tiny$\pm 0.10$ \textcolor{lgreen}{\small \ ($+25.6$)}  & $97.21   $\tiny$\pm 0.10$ \textcolor{lgreen}{\small \ ($+1.9$)}\vspace{3pt}\\
                                    & \multirow{2}{*}{SPLoRA-$r8$}  & 30\% & \phantom{$0{,}$}$205.3  $\tiny$\pm 0.1$ \textcolor{lgreen}{\small \ ($\downarrow 16.5\times$)}  & \phantom{$0{,}$}$566.2  $\tiny$\pm 10.9$  & $93.98  $\tiny$\pm 0.24$ \textcolor{lred}{\small \ ($-1.7$)}  & $91.51  $\tiny$\pm 0.49$ \textcolor{lred}{\small \ ($-1.4$)}  & $97.90  $\tiny$\pm 0.13$ \textcolor{lred}{\small \ ($-0.3$)}\\
                                    &                               & 10\% & \phantom{$0{,}$}$\mathbf{128.4}   $\tiny$\pm 0.1$ \textcolor{lgreen}{\small \ ($\downarrow 4.5\times$)}  & \phantom{$0{,}$}$243.2   $\tiny$\pm 9.8$  & $91.22   $\tiny$\pm 0.32$ \textcolor{lgreen}{\small \ ($+3.2$)}  & $86.76  $\tiny$\pm 0.42$ \textcolor{lgreen}{\small \ ($+21.1$)}  & $96.83   $\tiny$\pm 0.30$ \textcolor{lgreen}{\small \ ($+1.5$)}\\
\midrule
                                    & \multirow{2}{*}{Fine-pruning} & 30\% & $4{,}428.1                                         $\tiny$\pm 20.6$  & \phantom{$0{,}$}$719.9   $\tiny$\pm 0.7$  & $\mathbf{96.54}                               $\tiny$\pm 0.14$  & $\mathbf{95.37}                               $\tiny$\pm 0.08$  & $\mathbf{98.65}                               $\tiny$\pm 0.07$\\
                                    &                               & 10\% & \phantom{$0{,}$}$608.4                                          $\tiny$\pm 6.3$  & \phantom{$0{,}$}$301.6   $\tiny$\pm 2.3$  & $93.52                               $\tiny$\pm 0.05$  & $87.56                               $\tiny$\pm 2.75$  & $-$\vspace{3pt}\\
\multirow{2}{*}{LRP}           
                                    & \multirow{2}{*}{SPLoRA-$r32$} & 30\% & \phantom{$0{,}$}$592.6   $\tiny$\pm 0.9$ \textcolor{lgreen}{\small \ ($\downarrow 7.5\times$)}  & \phantom{$0{,}$}$585.5   $\tiny$\pm 6.7$  & $94.85  $\tiny$\pm 0.13$ \textcolor{lred}{\small \ ($-1.7$)}  & $93.62  $\tiny$\pm 0.38$ \textcolor{lred}{\small \ ($-1.8$)}  & $98.09  $\tiny$\pm 0.11$ \textcolor{lred}{\small \ ($-0.6$)}\\
                                    &                               & 10\% & \phantom{$0{,}$}$290.9   $\tiny$\pm 0.2$ \textcolor{lgreen}{\small \ ($\downarrow 2.1\times$)}  & \phantom{$0{,}$}$\mathbf{270.4}   $\tiny$\pm 6.4$  & $93.47   $\tiny$\pm 0.36$ \textcolor{lgreen}{\small \ ($+0.1$)}  & $91.22   $\tiny$\pm 0.46$ \textcolor{lgreen}{\small \ ($+3.7$)}  & $97.01  $\tiny$\pm 0.27$ 
                                    \vspace{3pt}\\
                                    & \multirow{2}{*}{SPLoRA-$r8$}  & 30\% & \phantom{$0{,}$}$203.3  $\tiny$\pm 0.5$ \textcolor{lgreen}{\small \ ($\downarrow 21.8\times$)}  & \phantom{$0{,}$}$591.1  $\tiny$\pm 12.4$  & $93.53  $\tiny$\pm 0.18$ \textcolor{lred}{\small \ ($-3.0$)}  & $91.26  $\tiny$\pm 0.19$ \textcolor{lred}{\small \ ($-4.1$)}  & $97.80  $\tiny$\pm 0.20$ \textcolor{lred}{\small \ ($-0.8$)}\\
                                    &                               & 10\% & \phantom{$0{,}$}$\mathbf{128.0}   $\tiny$\pm 0.1$ \textcolor{lgreen}{\small \ ($\downarrow 4.7\times$)}  & \phantom{$0{,}$}$281.6   $\tiny$\pm 1.7$  & $90.94  $\tiny$\pm 0.39$ \textcolor{lred}{\small \ ($-2.5$)}  & $85.69  $\tiny$\pm 1.39$ \textcolor{lred}{\small \ ($-1.9$)}  & $96.88  $\tiny$\pm 0.52$ 
                                    \\
\bottomrule 
\end{tabular}
}
\end{center}
\end{table*}

\subsection{Experimental setup}
In this set of experiments, we reuse and augment a previously reported setup~\cite{yeom2021pruning} to perform transfer-pruning for ResNet models pretrained on ILSVRC 2012~\cite{russakovsky2015imagenet} to the image classification benchmarks 
CIFAR-10~\cite{krizhevsky09learning}, Oxford Flowers 102~\cite{nilsback2008automated}, and Cats and Dogs~\cite{elson2007asirra}. 
As no publicly available test split was available for the latter,
we defined train-test splits and preprocessed data using DatasetOps~\cite{hedegaard2022datasetops} to match the 8005 training and 2023 test samples reported previously~\cite{yeom2021pruning}.
For transfer pruning, we first train the network without pruning for $\{30, 100, 100\}$ epochs 
and then perform incremental pruning at increments of 5\% until 5\% of weights remain in total; the incremental pruning is interspersed with $\{20, 50, 50\}$ epochs of training for the \{CIFAR-10, Oxford Flowers 102, Cats and Dogs\} datasets. \cref{apx:training-times} presents an overview of training times.
We employ the Strochastic Gradient Descent optimizer with a momentum of $0.9$, weight decay of $5 \cdot 10^{-4}$, and learning rate of $0.01$ at a batch size $256$ or down-scaled rates following the \textit{linear scaling rule}~\cite{krizhevsky2014estimating} when GPU memory limitations must be accomodated. In each training block, we use a step learning rate reduction of $5\times$ after each quarter of epochs. The above setup is used for either fine-pruning, in which all model weights are updated, or adaptation and pruning, which freezes the original network weights and only trains the adapter weights, normalization, and prediction head.

\begin{figure*}[tb]
    \centering
    \begin{subfigure}{0.245\linewidth}
        \centering
        \includegraphics[width=\linewidth]{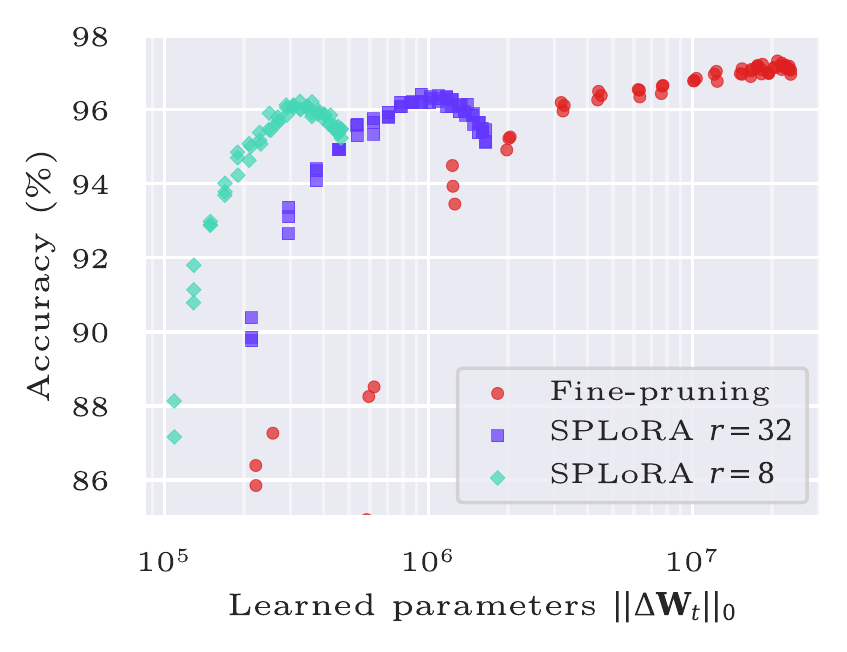}
        \\
        \includegraphics[width=\linewidth]{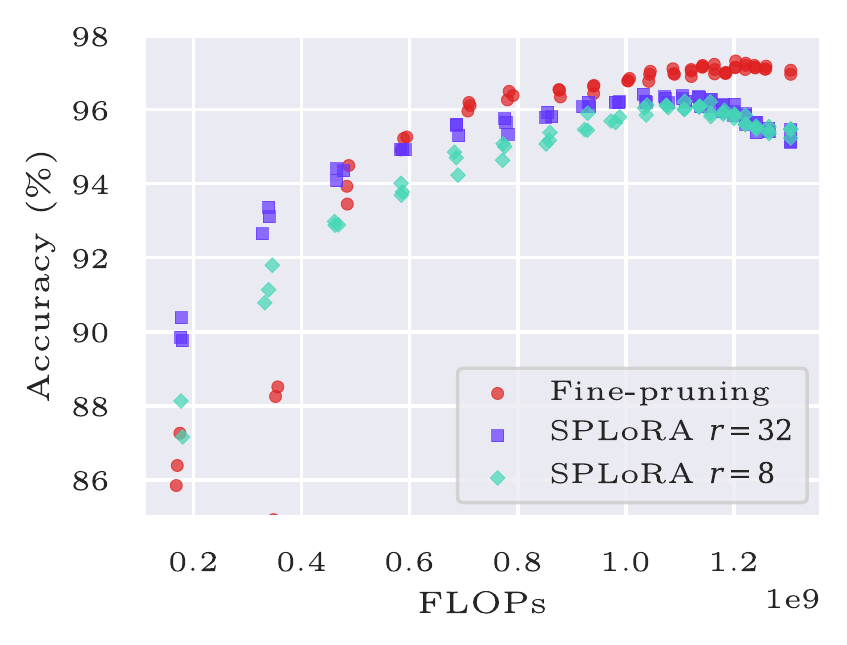}
        \caption{Weight pruning.}
        \label{fig:acc-params-weight}
    \end{subfigure}
    \hfill
    \begin{subfigure}{0.245\linewidth}
        \centering
        \includegraphics[width=\linewidth]{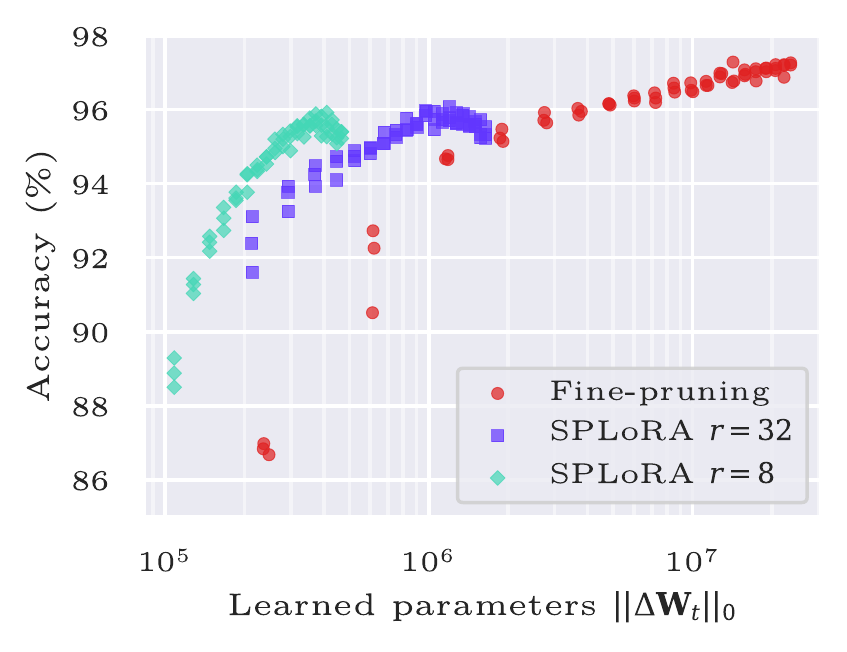}
        \\
        \includegraphics[width=\linewidth]{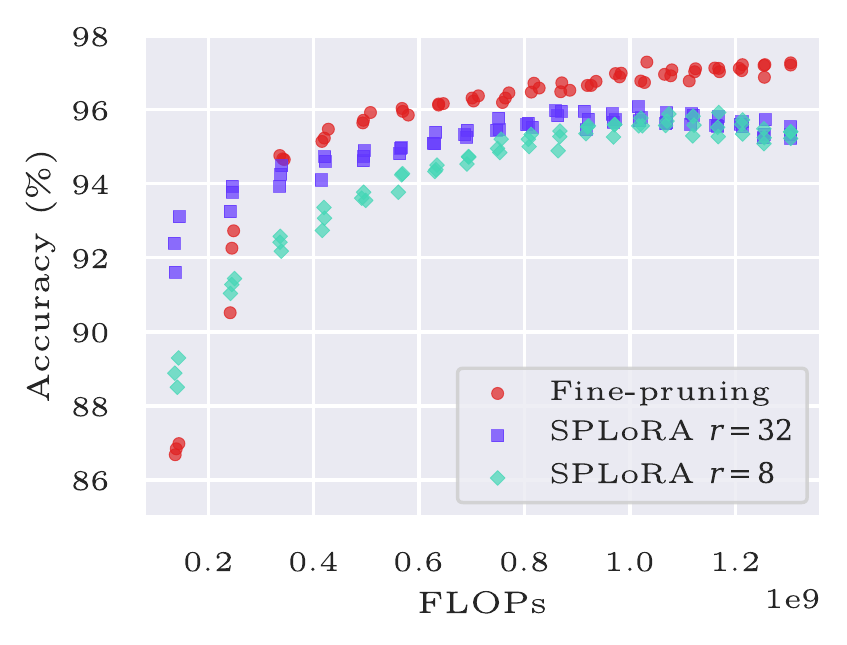}
        \caption{Gradient pruning.}
        \label{fig:acc-params-gradient}
    \end{subfigure}
    \hfill
    \begin{subfigure}{0.245\linewidth}
        \centering
        \includegraphics[width=\linewidth]{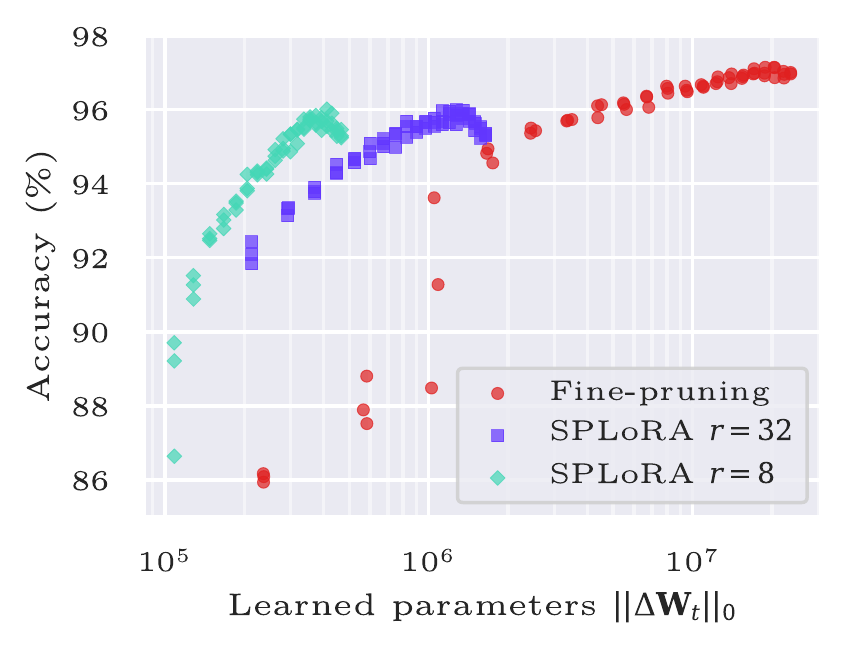}
        \\
        \includegraphics[width=\linewidth]{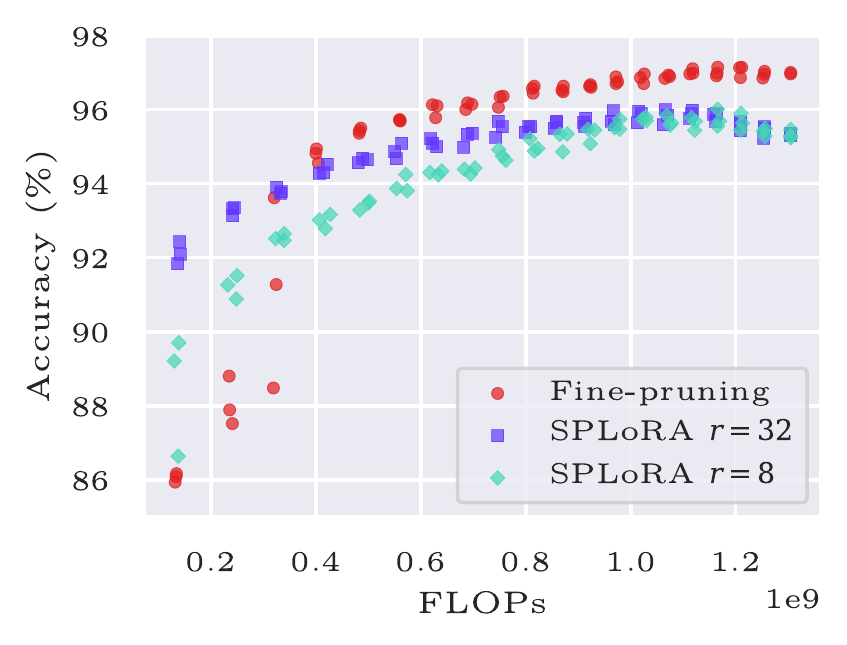}
        \caption{Taylor pruning.}
        \label{fig:acc-params-taylor}
    \end{subfigure}
    \hfill
    \begin{subfigure}{0.245\linewidth}
        \centering
        \includegraphics[width=\linewidth]{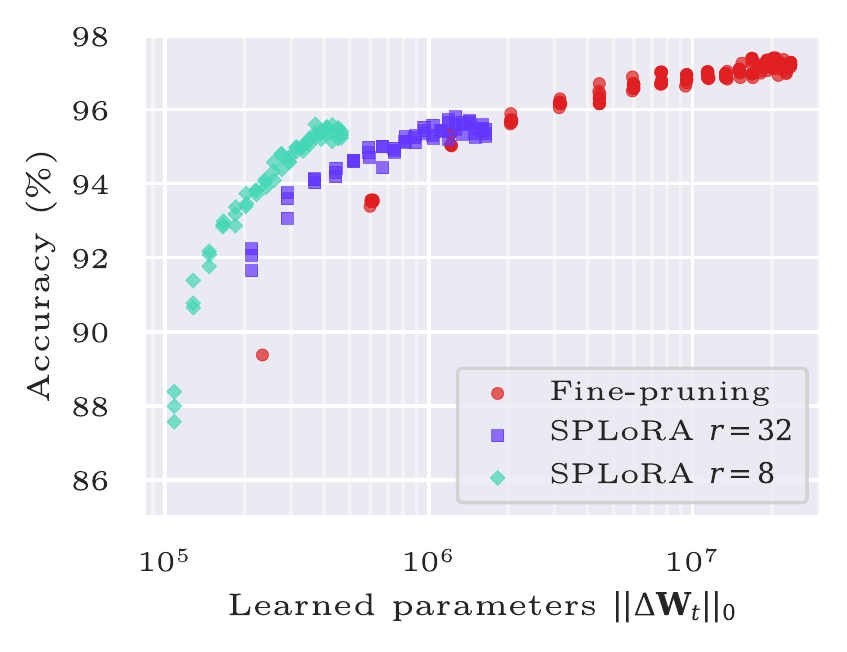}
        \\
        \includegraphics[width=\linewidth]{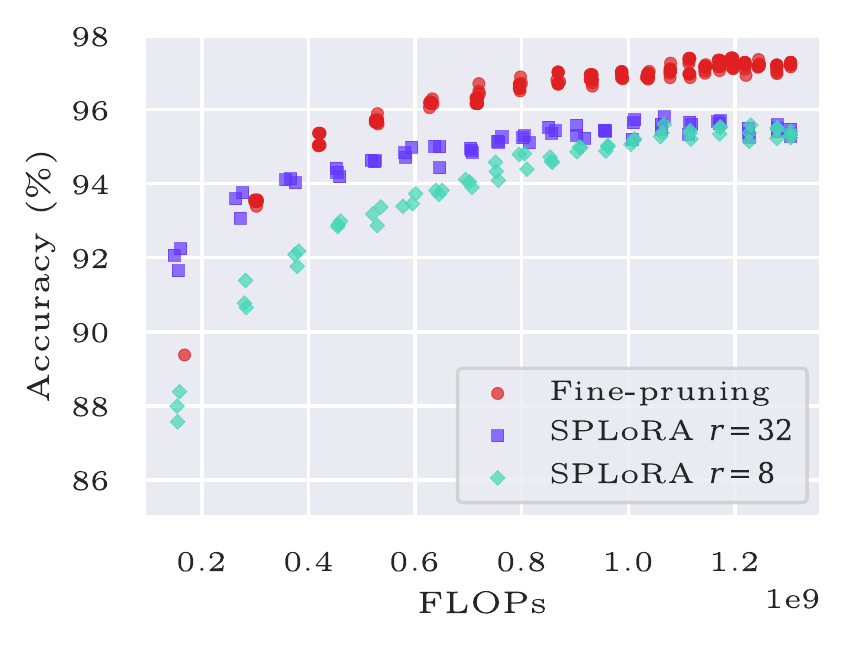}
        \caption{LRP pruning.}
        \label{fig:acc-params-lrp}
    \end{subfigure}
    \caption{CIFAR-10 accuracy versus total learned parameter count $\lVert \Delta\mW_t \rVert_0$ (top row) and FLOPs (bottom row) using fine-pruning and SPLoRA with ranks 32 and 8 for the (a) Weight~\cite{li2017pruning}, (b) Gradient~\cite{sun17meprop}, (c) Taylor~\cite{molchanov2017prining}, and (d) LRP~\cite{yeom2021pruning} channel-pruning methods.
    }
    \label{fig:cifar-10-results}
\end{figure*}

\begin{figure}[tb]
    \centering
    \includegraphics[width=\linewidth]{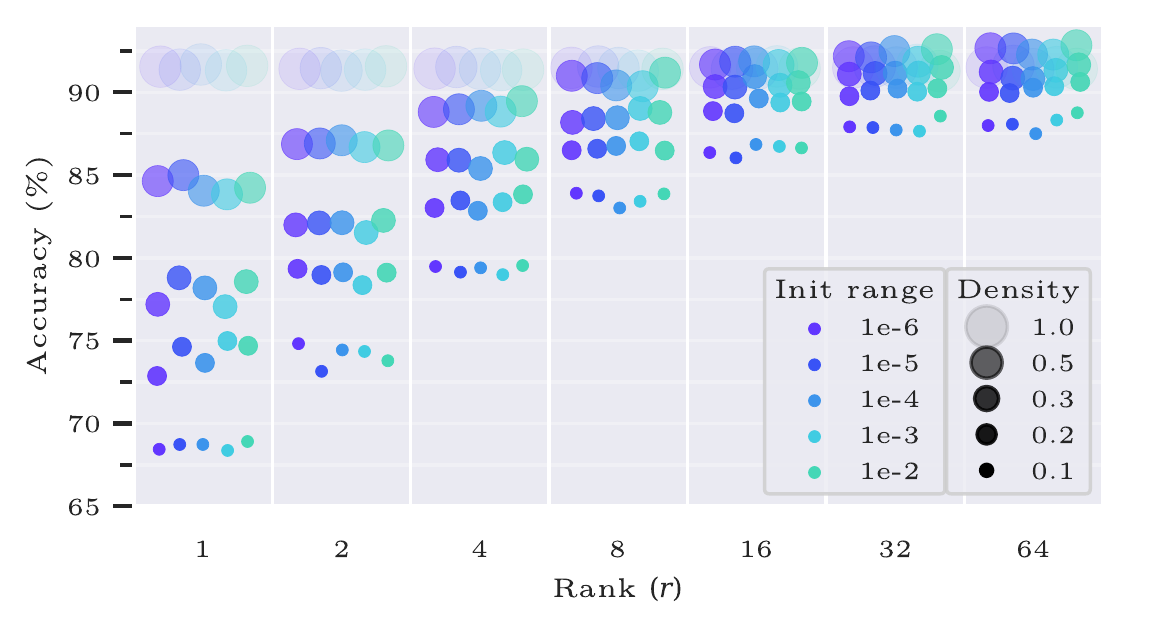}
    \caption{ResNet-18 accuracy on CIFAR-10 using SPLoRA of varying ranks ($r$) and adapter weight initialization ranges at different model densities ($d$). Point size $\propto d$ with transparency $\propto -d$.}
    \label{fig:SPLoRA-ablation-dots}
\end{figure}

\subsection{SPLoRA initialization and rank choice}\label{sec:splora-sensitivity}
To gauge the sensitivity of SPLoRA hyper-parameters and their effect on predictive performance, we perform a set of adaptation and pruning runs using $L_2$-normalized Taylor pruning~\cite{molchanov2017prining} on CIFAR-10 with a ResNet-18 network. Here, we vary the rank $r \in 2^{[0, 6]}$ and initialization range in $10^{[-6, -2]}$ and evaluate along densities $\{1.0, 0.5, 0.3, 0.2, 0.1\}$.
As illustrated in \cref{fig:SPLoRA-ablation-dots}, we observe a clear and expected trend of increasing accuracy as the rank is increased. The increases exhibit diminishing returns with limited benefit beyond $r=32$ for CIFAR-10. 
While all tested ranks show similar accuracy at a density of $d=1.0$, the lowest-rank adapters are more severely affected by a lower $d$ than higher-rank ones.
This follows intuition, considering that lower-rank adapters have fewer parameters that might prove redundant during pruning.
In some cases ($r \ge 16$) pruning at $d\approx0.5$ increases accuracy. 
SPLoRA is generally robust to the chosen initialization range, showing no clear trends in favor of particular ranges. We will use $\mW_\text{up}^{(i,j)} \sim \mathcal{U}(-10^{-4}, 10^{-4})$ in subsequent experiments.

\subsection{Comparison with fine-pruning} \label{sec:experiments-splora-comp-with-fineprune}
In our comparison of SPLoRA and fine-pruning, we train a ResNet-50 network for weight-based transfer learning from ILSVRC 2022 to CIFAR-10, Oxford Flower 102, and Cats and Dogs across four structured pruning works, namely, the normalized Weight~\cite{li2017pruning}, Taylor~\cite{molchanov2017prining}, Gradient~\cite{sun17meprop}, and LRP~\cite{li2017pruning} pruning methods. This comparison is conducted for both fine-pruning and SPLoRA with the ranks $r=8$ and $r=32$.
To accommodate the stochastic nature of pruning and neural network training, we repeat each experiment three times and report the mean and standard deviation of each metric.
The results of our experiments are presented in \Cref{tab:resnet50} for model densities of $100\%$, $30\%$, and $10\%$ retained weights and visualized for CIFAR-10 in \cref{fig:cifar-10-results} as well as for Oxford Flowers 102 and Cats and Dogs\footnote{We found LRP with fine-pruning to be unstable at low model densities. Despite attempts with multiple different seeds, results for Cats and Dogs could not be obtained (denoted by ``-'' in \Cref{tab:resnet50}).} in \cref{apx:splora-vis}

Comparing SPLoRA with fine-pruning, we observed competitive transfer adaptations despite learning a fraction the weights. While fine-pruning generally resulted in higher accuracy at 30\% model density (on average $0.6\%$ and $1.6\%$ higher than SPLoRA ranks 32 and 16), SPLoRA had far fewer learned parameters (on average $6.2\times$ and $17.0\times$). At 10\% density, SPLoRA was both more robust to pruning (achieving $6.9\%$ and $4.7\%$ higher average accuracy than fine-pruning for ranks 32 and 16), while reducing the number of learned parameters (by $2.0\times$ and $4.2\times$ on average). 
The difference in parameter reduction between the 30\% and 10\% density targets is explained partly by the normalization and linear parameters (73K for ResNet-50), that begin to dominate the learned weight budget at low densities, and partly by the remaining weights of adapter bottleneck channels. Even fewer parameters would be needed if a lower rank hyper-parameter was chosen or if, as discussed in \cref{sec:limitations}, pruning of bottleneck channels was conducted as well.
As FLOPs follow model densities, these are approximately equal for each learning method given equal densities.

\section{Conclusion}\label{sec:conclusion}
We proposed Structured Pruning Adapters (SPAs) as an alternative to fine-tuning during structured pruning.
Instead of updating all model weights, SPAs consist of prunable lightweight add-on modules, which are learned in place of the original weights but can be fused with them at run-time to obtain the same computational enhancements as regular structured pruning with fine-tuning. 
Our channel-based SPAs were shown to achieve competitive performance across a battery of pruning methods on computer vision benchmarks while requiring a fraction of the learned parameters per task. Thus, SPAs are ideal for task-switching storage-constrained and/or network-limited usage scenarios, where the per-model size should be small. 

\subsection*{Acknowledgments}
Lukas Hedegaard and Alexandros Iosifidis acknowledge funding from the European Union’s Horizon 2020 research and innovation programme under grant agreement No 871449 (OpenDR).

An acknowledgment also goes to Martin Damgaard Nielsen and Jens Dalgaard Nielsen for the early-stage conversations relating to this project.


\bibliography{bibliography}
\bibliographystyle{icml2023}

\newpage
\appendix
\onecolumn
\section{SPLoRA Derivation} \label{apx:smola-derivation}
Utilizing \cref{eq:mlora} in the context of channel-SPAs of the form expressed in \cref{eq:channel-parallel}, we can derive the Structured Pruning Low-rank Adapter defined in \cref{eq:splora}, where adapter parameters are pruned alongside source weights. 
This derivation is straightforward, considering that the application of a structured pruning mask $\mM = \vm_\text{row} \vm_\text{col}^\top$ via Hadamard products is equivalent to a projection with diagonalized masking vectors:
\begin{equation}
    \mW \odot \vm_\text{row} \vm_\text{col}^\top
    = \text{diag}(\vm_\text{row}) \ \mW \ \text{diag}(\vm_\text{col}).
    \label{eq:masking-identity}
\end{equation}
Similarly, a single diagonalized mask can be expressed via Hadamark products:
\begin{equation}
    \text{diag}(\vm_\text{row}) \ \mW  
    =
    \mW \odot \vm_\text{row} \vone^\top.
    \label{eq:masking-identity-single}
\end{equation}
Utilizing \cref{eq:mlora}, \cref{eq:masking-identity}, and \cref{eq:masking-identity-single}, we can rewrite \cref{eq:channel-parallel} as follows:
\begin{align*}
    \mW_t
    &= (\mW_s + a(\Delta \mW_t)) \odot \vm_\text{row} \vm_\text{col}^\top \\
    &= (\mW_s + \mW_\text{down}\mW_\text{up}) \odot \vm_\text{row} \vm_\text{col}^\top \\
    &= \text{diag}(\vm_\text{row}) \ (\mW_s + \mW_\text{down}\mW_\text{up}) \ \text{diag}(\vm_\text{col}) \\
    &= \text{diag}(\vm_\text{row})\ \mW_s \text{diag}(\vm_\text{col}) 
        + (\text{diag}(\vm_\text{row}) \mW_\text{down})(\mW_\text{up} \text{diag}(\vm_\text{col})) \\
    &= \mW_s \odot \vm_\text{row} \vm_\text{col}^\top 
        + (\mW_\text{down} \odot \vm_\text{row} \vone^\top ) (\mW_\text{up} \odot \vone \vm_\text{col}^\top ),
\end{align*}
where the final result is equivalent to \cref{eq:splora}.
%

\section{Training Durations} \label{apx:training-times}
In this section, we provide a brief overview of approximate training durations for the methods tested in the present paper.
As training times are comparable among different pruning methods, we report a single metric approximated from multiple pruning methods. These are presented in \Cref{tab:training-times-cv} for our experiments using ResNet-50 in image recognition tasks. 
Here, the pruning methods gradually reduce the network density while producing pruned models at a predefined step reduction in density, cycling the learning rate for each density reduction step. Accordingly, the noted training times for the pruned learning methods in \Cref{tab:training-times-cv} includes the training of all models with densities ranging from 100\% to 5\% at 5\% intervals.


\begin{table}[h!]
\caption{Training durations for the ResNet-50 model on image recognition transfer tasks using a NVIDIA RTX 2080 Ti GPU. For each dataset, the batch size (BS) and training duration (T) are presented.}
\label{tab:training-times-cv}
\begin{center}
\begin{tabular}{llcccccc}
\toprule 
\multirow{1}{*}{\bf Pruning}
    & \multirow{1}{*}{\bf Learning}
    &\multicolumn{2}{c}{\bf CIFAR-10}
    &\multicolumn{2}{c}{\bf Oxf. Fl. 102}
    &\multicolumn{2}{c}{\bf C. \& D.}
\\ 
{\bf method}  
    & {\bf method}
    & BS
    & T
    & BS
    & T
    & BS
    & T
    \\
\midrule 
\multirow{1}{*}{Unpruned}   & Fine-tuning   & 64 & 0:30h & 64 & 1:05h & 64 & 2:20h \\
                            & SPLoRA-$r32$  & 64 & 0:30h & 32 & 1:10h & 32 & 2:30h \\
                            & SPLoRA-$r8$   & 64 & 0:30h & 32 & 1:10h & 32 & 2:30h \\
\midrule 
\multirow{1}{*}{Pruned }    & Fine-pruning  & 64 & 10h & 4 & 25h & 6 & 42h \\
                            & SPLoRA-$r32$  & 64 & 11h & 4 & 33h & 6 & 48h \\
                            & SPLoRA-$r8$   & 64 & 11h & 4 & 31h & 6 & 45h \\
\bottomrule 
\end{tabular}
\end{center}
\end{table}





\section{Supplemental Visualisations of experimental results with SPLoRA} \label{apx:splora-vis}

\begin{figure*}[tb]
    \centering
    \begin{subfigure}{0.245\linewidth}
        \centering
        \includegraphics[width=\linewidth]{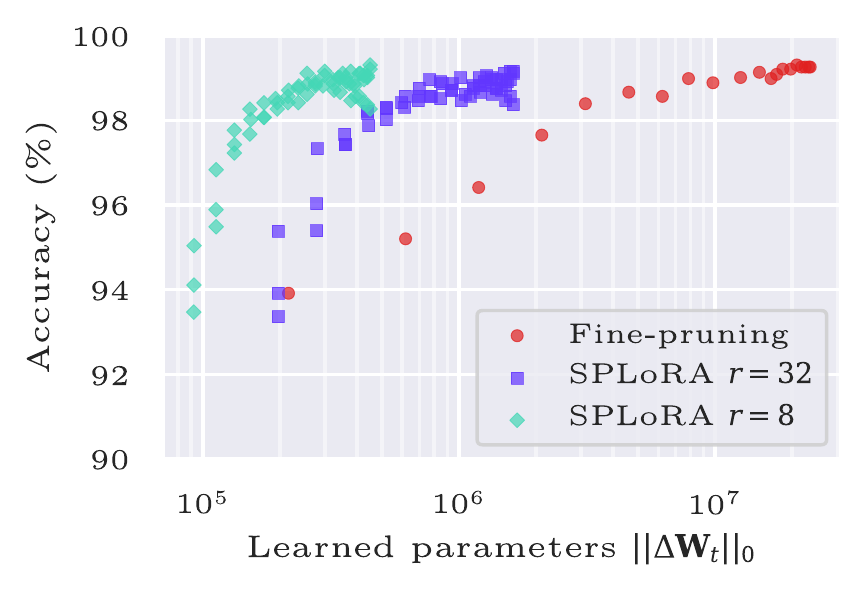}
        \\
        \includegraphics[width=\linewidth]{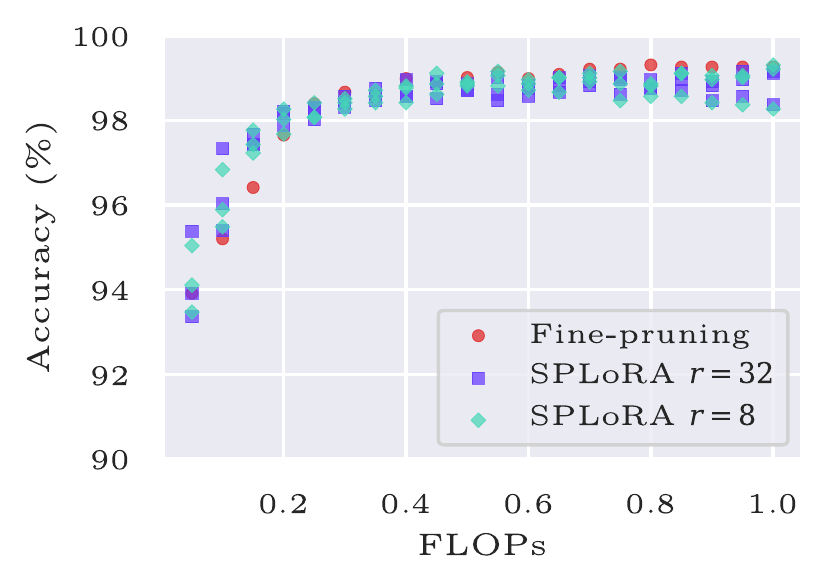}
        \
        \includegraphics[width=\linewidth]{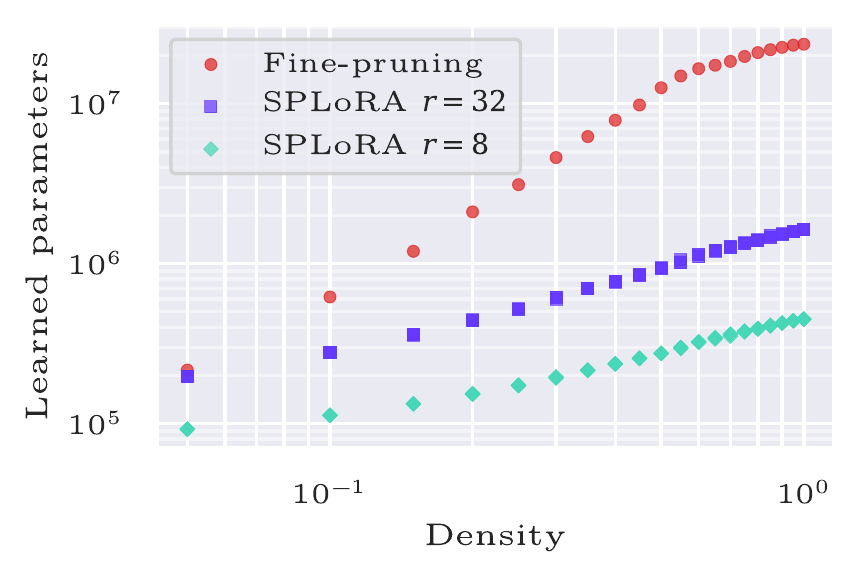}
        \caption{Weight pruning\\\cite{li2017pruning}.}
    \end{subfigure}
    \hfill
    \begin{subfigure}{0.245\linewidth}
        \centering
        \includegraphics[width=\linewidth]{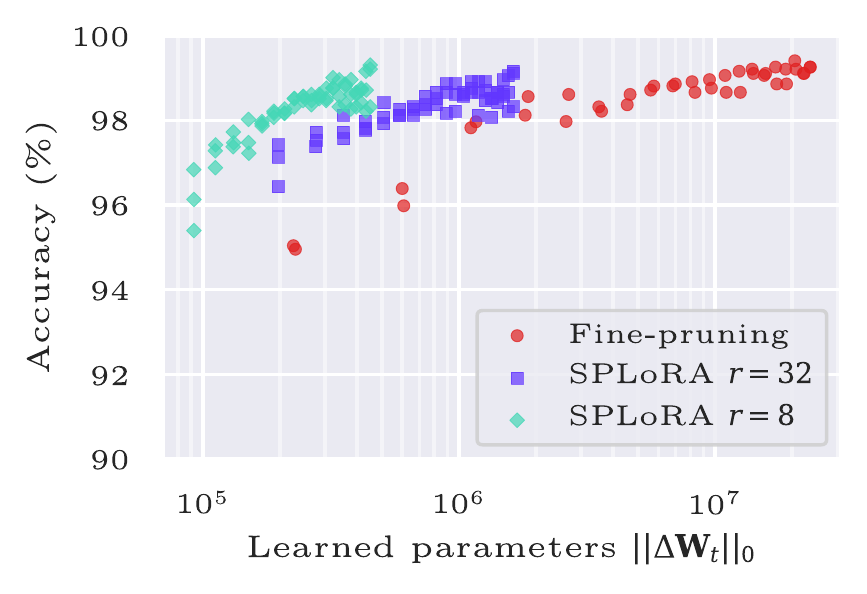}
        \\
        \includegraphics[width=\linewidth]{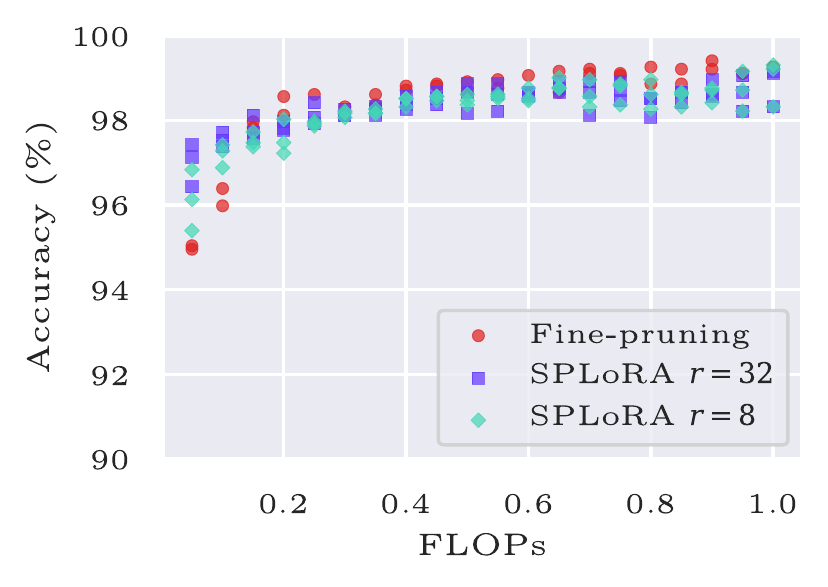}
        \\
        \includegraphics[width=\linewidth]{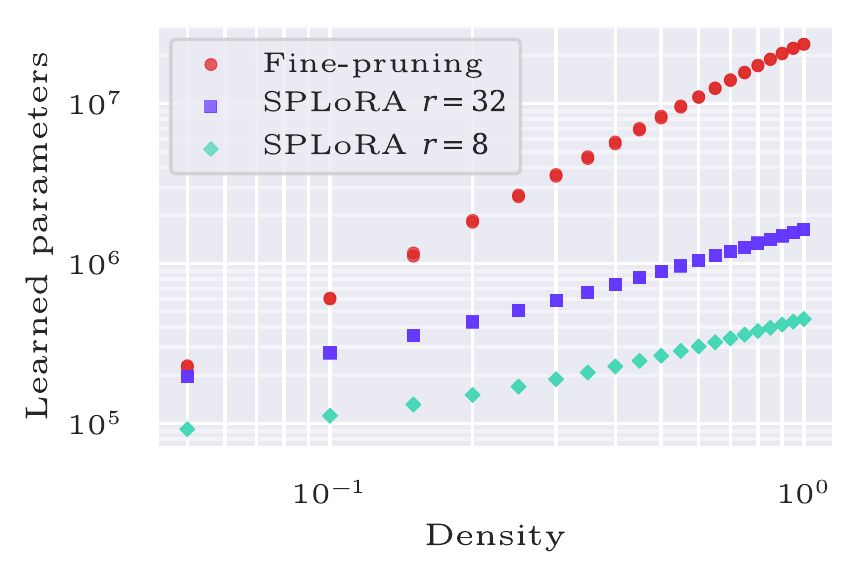}
        \caption{Gradient pruning\\\cite{sun17meprop}.}
    \end{subfigure}
    \hfill
    \begin{subfigure}{0.245\linewidth}
        \centering
        \includegraphics[width=\linewidth]{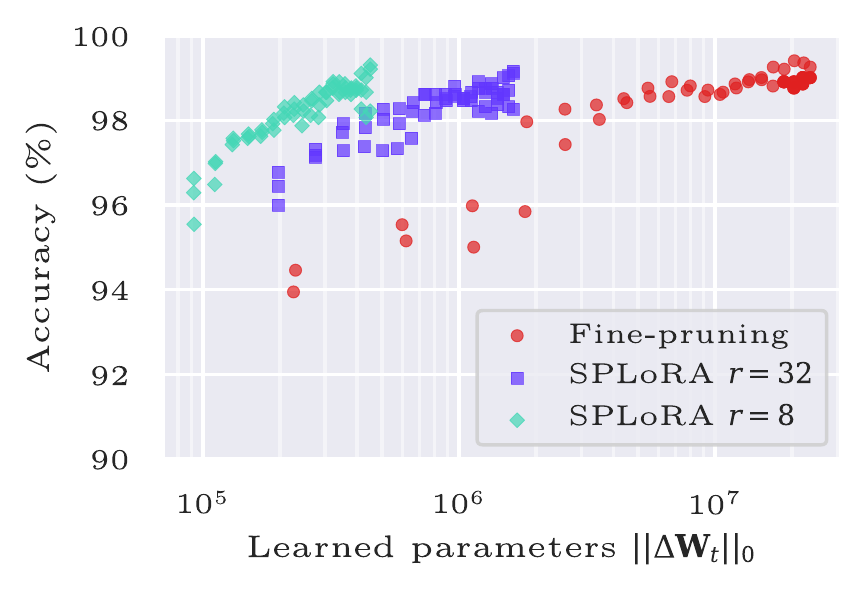}
        \\
        \includegraphics[width=\linewidth]{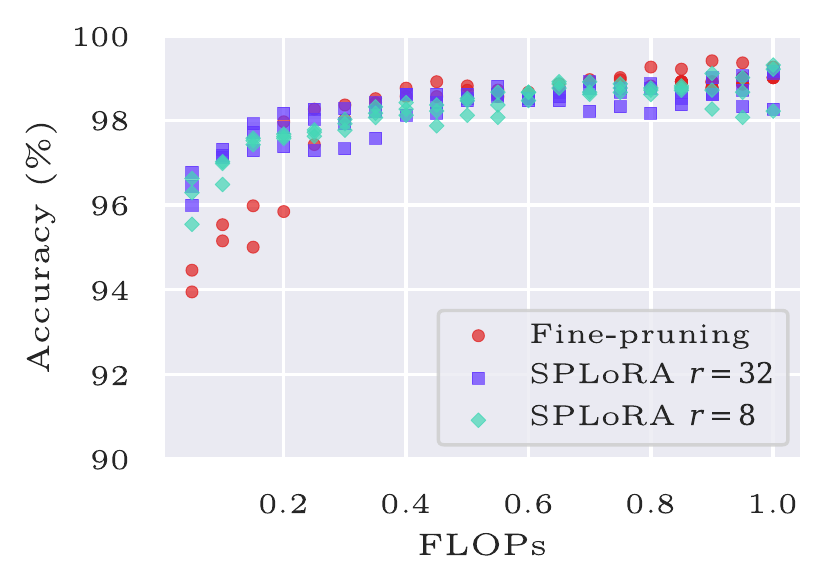}
        \\
        \includegraphics[width=\linewidth]{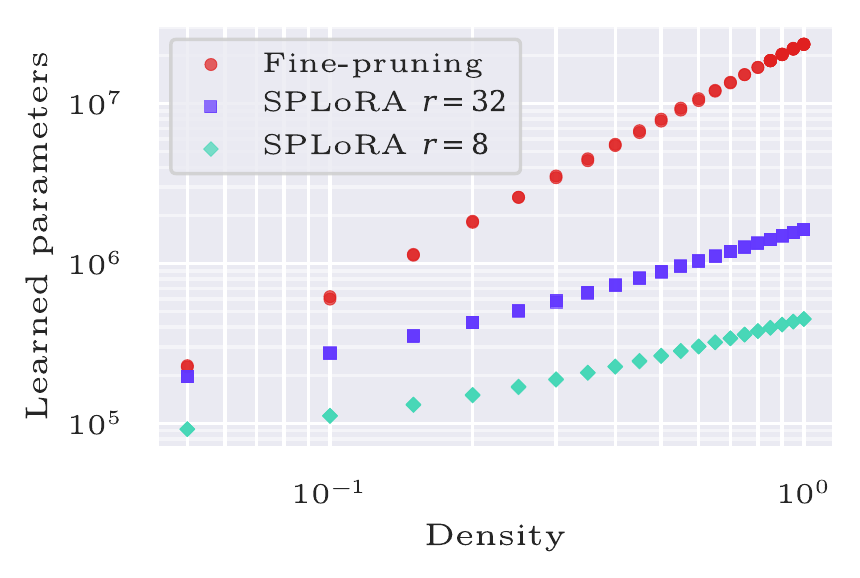}
        \caption{Taylor pruning\\\cite{molchanov2017prining}.}
    \end{subfigure}
    \hfill
    \begin{subfigure}{0.245\linewidth}
        \centering
        \includegraphics[width=\linewidth]{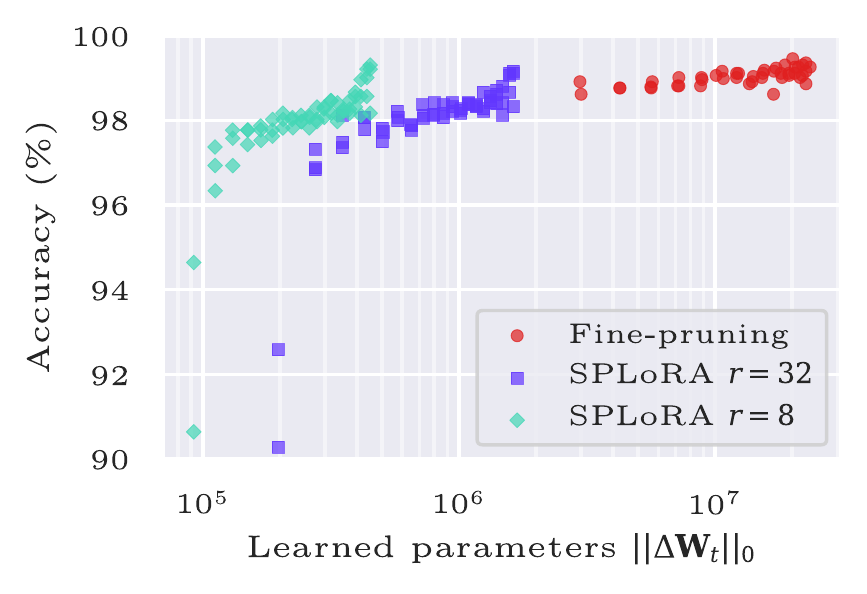}
        \\
        \includegraphics[width=\linewidth]{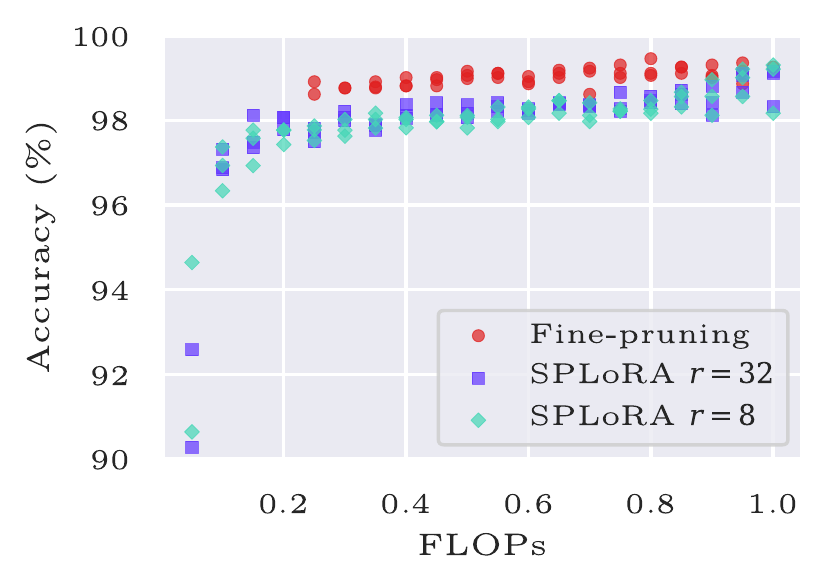}
        \\
        \includegraphics[width=\linewidth]{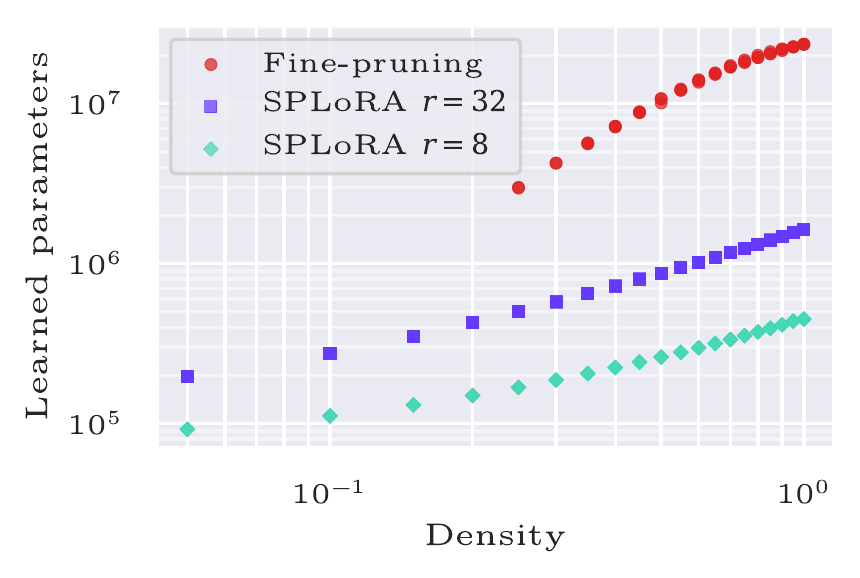}
        \caption{LRP pruning\\\cite{yeom2021pruning}.}
    \end{subfigure}
    \caption{Cats and Dogs accuracy versus learned parameter count $\lVert \Delta\mW_t \rVert_0$ (top row) and FLOPs (middle row) as well as learned parameter count versus model density (bottom row) using fine-pruning and SPLoRA with ranks 32 and 8 for various channel-pruning methods.
    }
    \label{fig:cats-and-dogs-curves}
\end{figure*}

\begin{figure*}[tb]
    \centering
    \begin{subfigure}{0.245\linewidth}
        \centering
        \includegraphics[width=\linewidth]{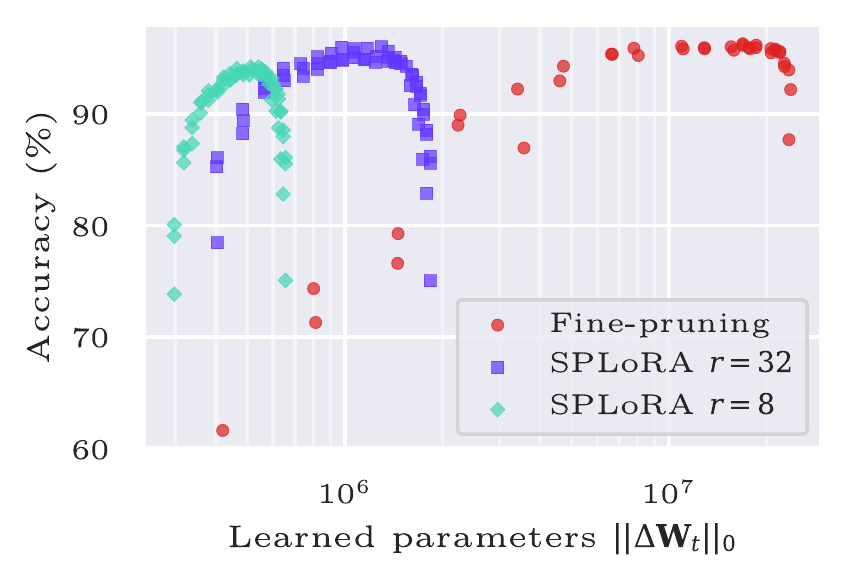}
        \\
        \includegraphics[width=\linewidth]{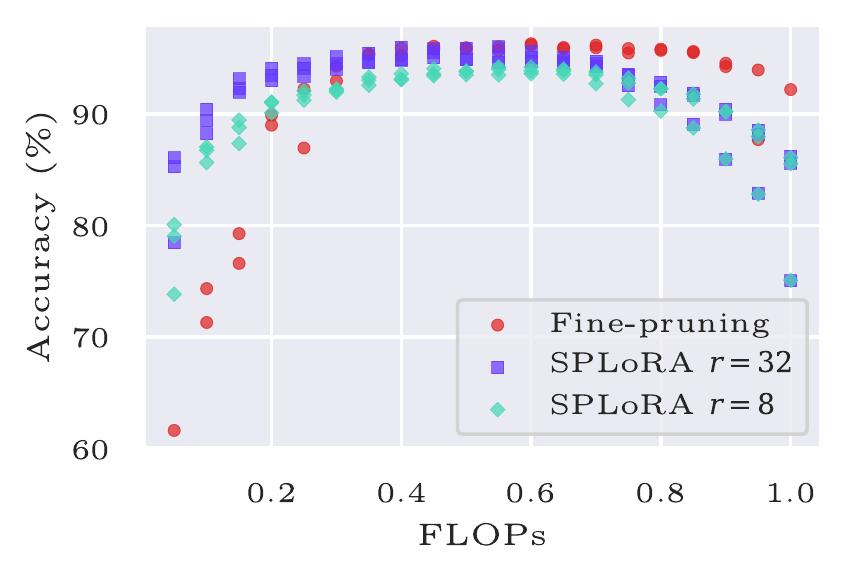}
        \
        \includegraphics[width=\linewidth]{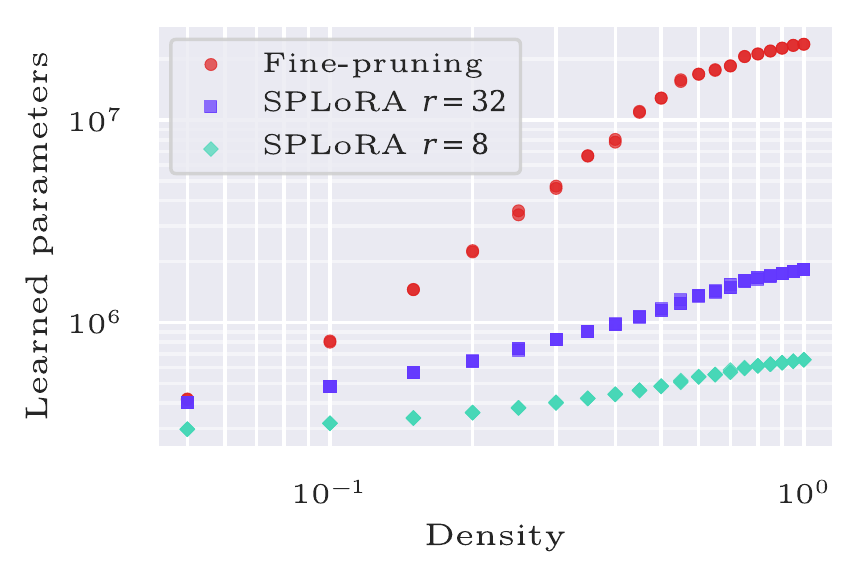}
        \caption{Weight pruning\\\cite{li2017pruning}.}
    \end{subfigure}
    \hfill
    \begin{subfigure}{0.245\linewidth}
        \centering
        \includegraphics[width=\linewidth]{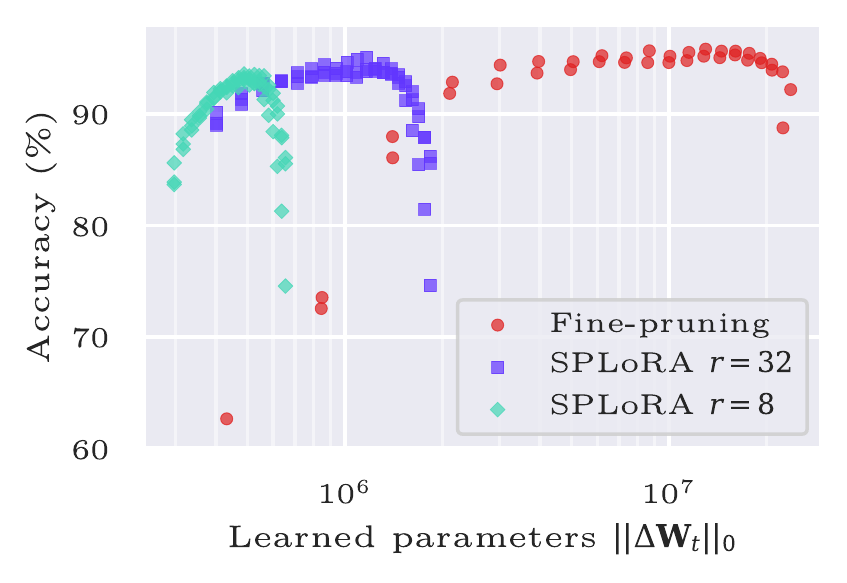}
        \\
        \includegraphics[width=\linewidth]{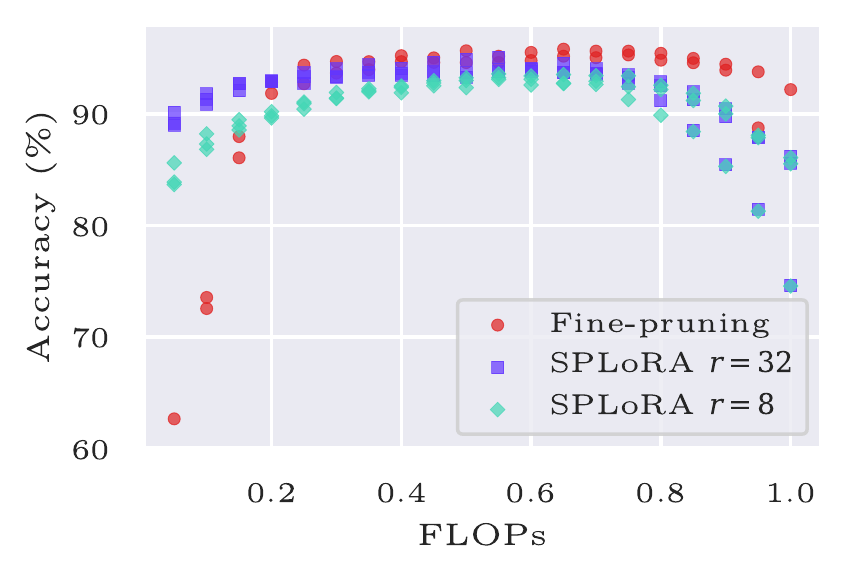}
        \\
        \includegraphics[width=\linewidth]{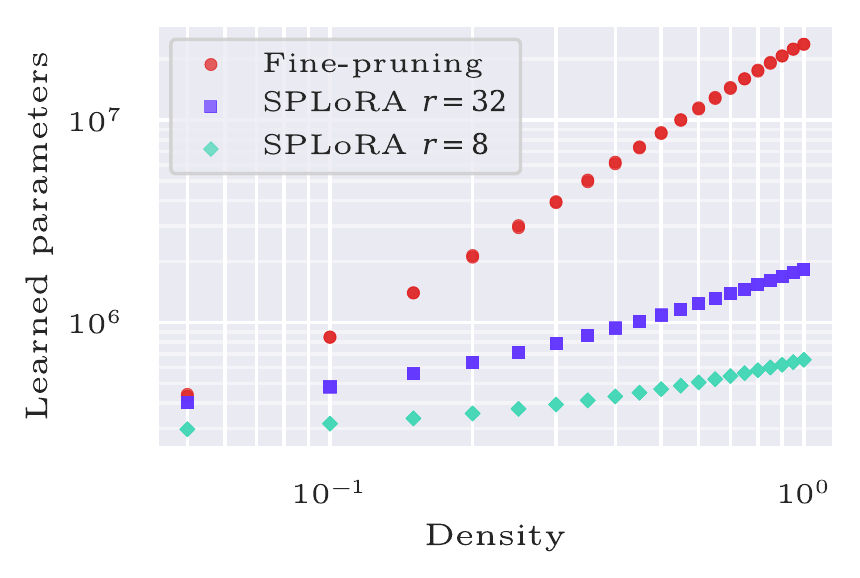}
        \caption{Gradient pruning\\\cite{sun17meprop}.}
    \end{subfigure}
    \hfill
    \begin{subfigure}{0.245\linewidth}
        \centering
        \includegraphics[width=\linewidth]{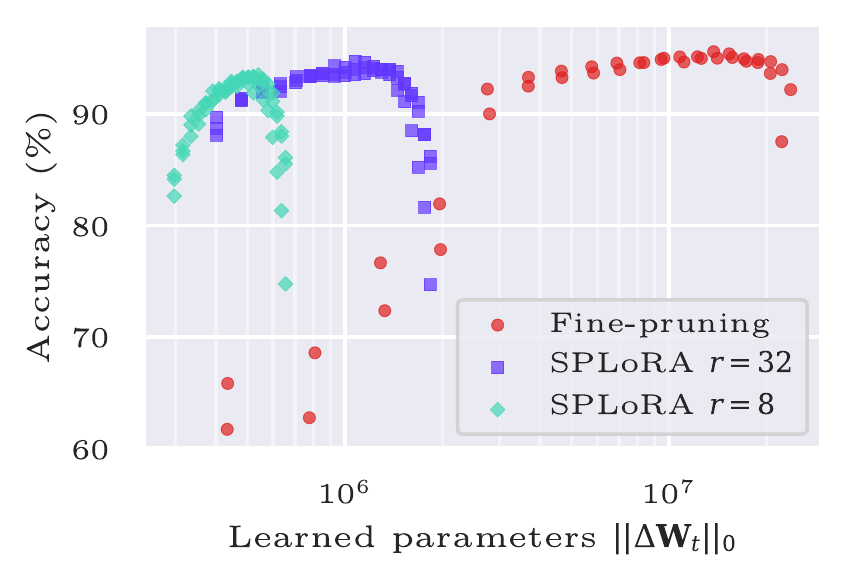}
        \\
        \includegraphics[width=\linewidth]{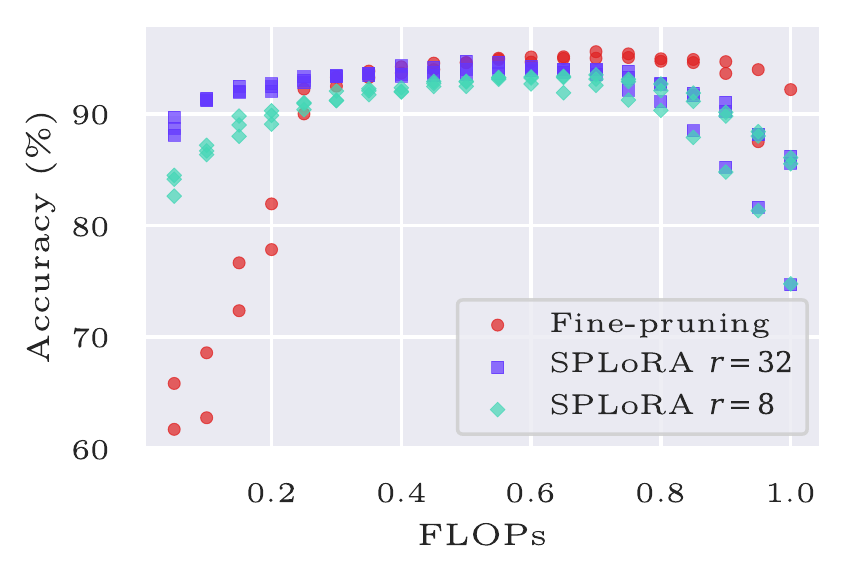}
        \\
        \includegraphics[width=\linewidth]{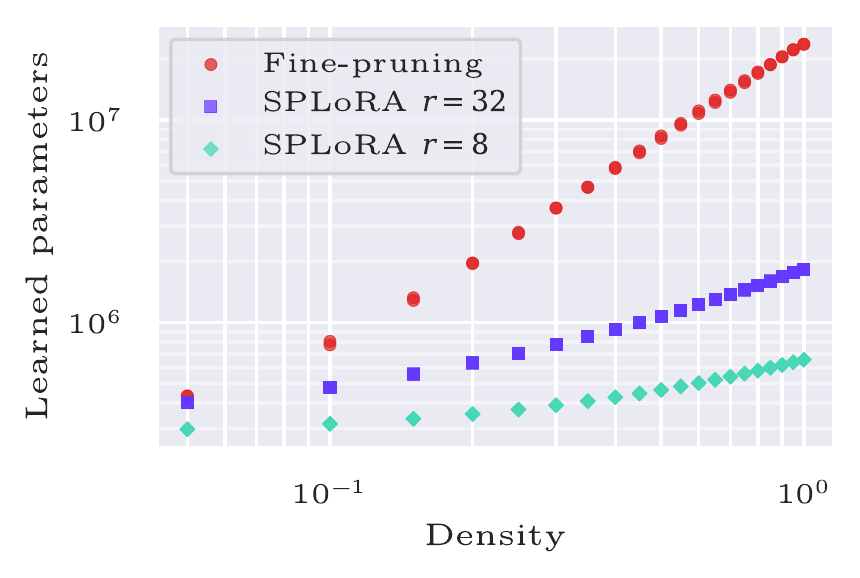}
        \caption{Taylor pruning\\\cite{molchanov2017prining}.}
    \end{subfigure}
    \hfill
    \begin{subfigure}{0.245\linewidth}
        \centering
        \includegraphics[width=\linewidth]{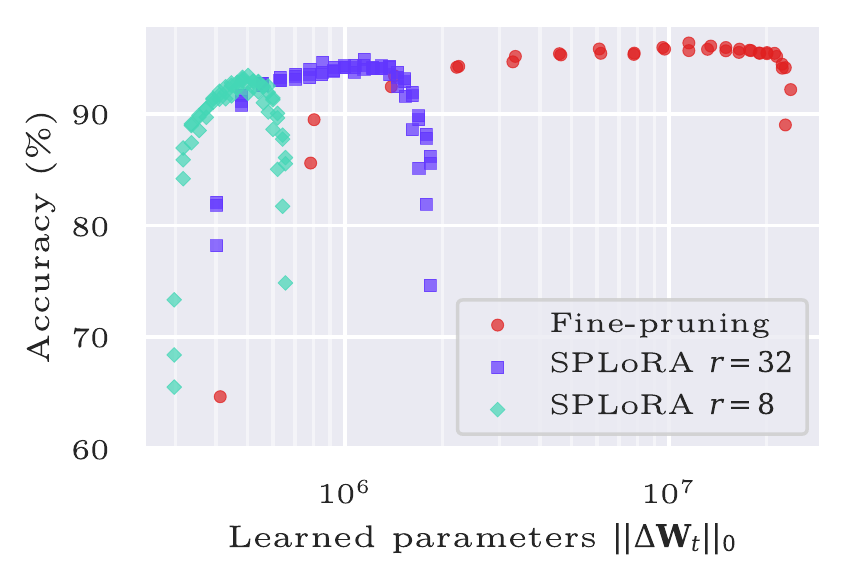}
        \\
        \includegraphics[width=\linewidth]{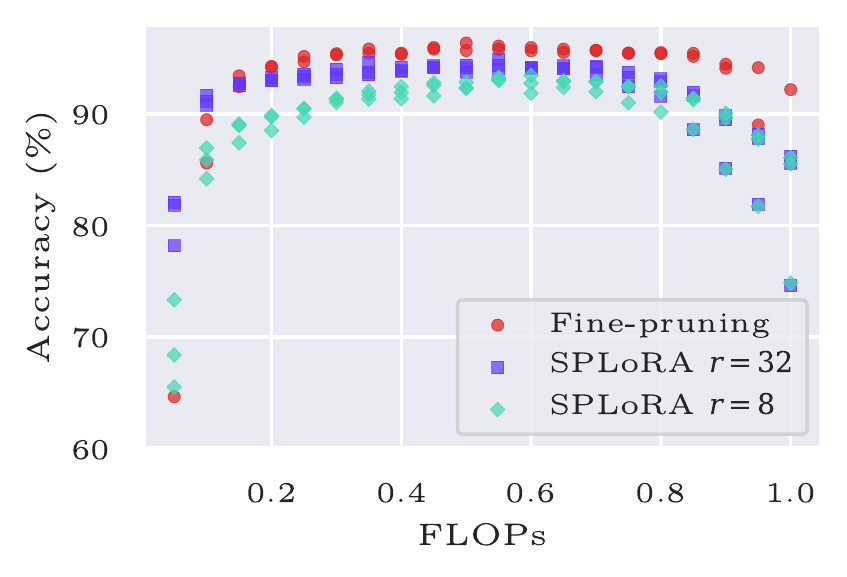}
        \\
        \includegraphics[width=\linewidth]{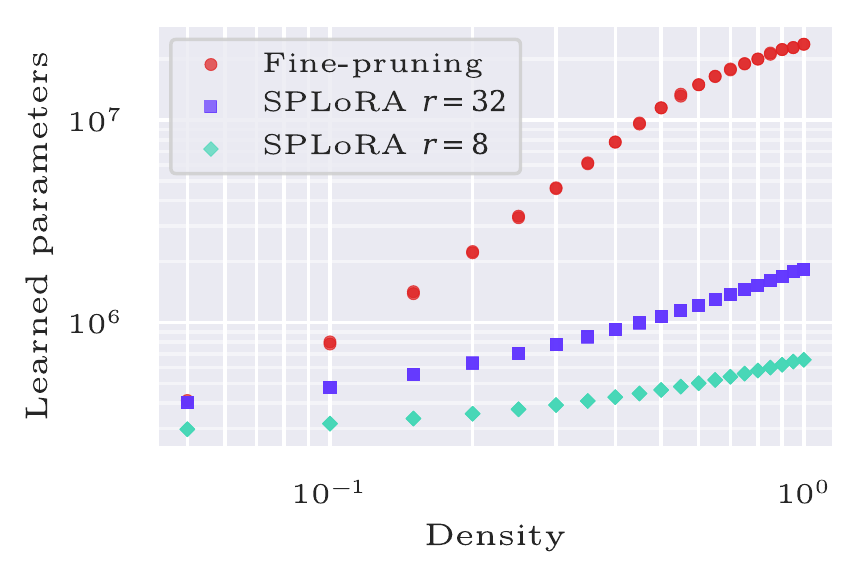}
        \caption{LRP pruning\\\cite{yeom2021pruning}.}
    \end{subfigure}
    \caption{Oxford Flowers 102 accuracy versus learned parameter count $\lVert \Delta\mW_t \rVert_0$ (top row) and FLOPs (middle row) as well as learned parameter count versus model density (bottom row) using fine-pruning and SPLoRA with ranks 32 and 8 for various channel-pruning methods.
    }
    \label{fig:oxford-flowers-curves}
\end{figure*}

For completeness, \cref{fig:cats-and-dogs-curves} and \cref{fig:oxford-flowers-curves} illustrate trade-offs among accuracy, learned parameters, and FLOPs.

\end{document}